\documentclass[11pt]{article}

\usepackage[preprint]{acl} % only for preprint

\usepackage{times}
\usepackage{latexsym}

\usepackage{hyperref}
\usepackage{url}
\usepackage{wrapfig}
\usepackage{booktabs}
\usepackage{subcaption}
\usepackage{graphicx}
\usepackage{multirow}
\usepackage{float}
\usepackage{algorithm}
\usepackage{algorithmic}
\usepackage{amsmath,amsfonts,bm}
\usepackage{amssymb}

\usepackage{array}
\usepackage[T1]{fontenc}
\usepackage[utf8]{inputenc}

\usepackage{microtype}

\usepackage{inconsolata}

\usepackage{graphicx}

\title{When Confidence Misleads: Suffix Anchoring and Anchor-Proximity Confidence Modulation for Diffusion Language Models}

\author{
 \textbf{Jungwon Park$^{1,5}$},
 \textbf{Jimyeong Kim$^{2}$},
 \textbf{Jungmin Ko$^{3}$},
 \textbf{Nojun Kwak$^{3,4}$},
 \textbf{Wonjong Rhee$^{3,4}$}
\\
 $^1$RICS, $^2$AIIS, $^3$IPAI, $^4$Department of Intelligence and Information, 
\\
 Seoul National University
\\ 
 $^5$Daegu Gyeongbuk Institute of Science and Technology 
\\
 \small{ \texttt{\{quoded97, wlaud1001, jungminko, nojunk, wrhee\}@snu.ac.kr}}
}

\begin{document}
\addtocontents{toc}{\protect\setcounter{tocdepth}{-1}}
\maketitle
% \begin{abstract}
% This document is a supplement to the general instructions for *ACL authors. It contains instructions for using the \LaTeX{} style files for ACL conferences.
% The document itself conforms to its own specifications, and is therefore an example of what your manuscript should look like.
% These instructions should be used both for papers submitted for review and for final versions of accepted papers.
% \end{abstract}

\begin{abstract} 
Diffusion language models generate text by iteratively selecting and denoising masked positions, making position selection a central inference-time decision. Most training-free methods rely on model confidence, assuming that high-confidence positions are ready to be decoded. However, this assumption can fail in fully non-autoregressive~(fully non-AR) decoding, where selection spans the entire masked response sequence. This can cause incomplete generation, partly due to EOT overconfidence. Existing remedies directly suppress EOT tokens, but provide limited gains or require additional fine-tuning. We show that placing even a semantically minimal suffix anchor near the end of the response region discourages premature EOT prediction and promotes complete generation. However, despite providing little supporting context, the anchor makes nearby positions highly confident and causes them to be decoded too early. We therefore propose \textbf{Suffix-Anchored Confidence Modulation}, a simple training-free method that uses suffix anchoring with progress-dependent confidence modulation near the anchor. Across text-only reasoning, vision-language reasoning, and code-generation benchmarks, our method outperforms diverse fully non-AR and semi-AR baselines. It also achieves particularly large gains under highly parallel decoding, demonstrating the benefit of improving the fully non-AR regime over block-wise semi-AR regime.
\end{abstract}
% Diffusion language models decode text by iteratively denoising masked token sequences, making the choice of which positions to decode a central inference-time decision. Most training-free decoding strategies use model confidence for position selection, assuming that high-confidence positions are ready to be decoded. In this work, we revisit this assumption by studying when confidence misleads fully non-autoregressive~(fully non-AR) decoding. EOT tokens can receive high confidence and cause incomplete generation; inserting a suffix anchor can mitigate this issue but introduces local overconfidence near the anchor, causing anchor-adjacent tokens to be decoded too early. To address these issues, we propose \textbf{Suffix-Anchored Confidence Modulation}, a simple training-free method that inserts a short suffix anchor to encourage response completion and modulates confidence near the anchor according to decoding progress. This preserves the response-completion benefit of suffix anchoring while reducing premature decoding of anchor-adjacent tokens. Across text-only reasoning, vision-language reasoning, and code-generation benchmarks, our method consistently improves confidence-based fully non-AR decoding, outperforms explicit EOT suppression, and preserves the parallel decoding advantage of fully non-AR generation.
\section{Introduction}
\label{sec:introduction}

Diffusion Language Models~(DLMs) generate text by iteratively selecting and denoising masked positions, allowing multiple tokens to be decoded in parallel rather than strictly generating tokens from left to right~\citep{lou2023discrete, shi2024simplified, sahoo2024simple, zheng2025masked, ou2025your}. This makes position selection a central inference-time decision: at each decoding step, the model must determine not only what tokens to predict, but also which masked positions to unmask.

Most training-free DLM decoding strategies use model confidence for position selection. For example, \textit{top-probability decoding}~\citep{chang2022maskgit, nie2026large}, also known as \textit{low-confidence remasking}, selects positions whose predicted tokens have the highest probability, while \textit{top-margin decoding}~\citep{kim2025train} favors positions whose top two predictions are well separated. These strategies implicitly assume that high-confidence positions are ready to be decoded. However, this assumption can fail in fully non-autoregressive~(fully non-AR) decoding, where position selection spans the entire masked response sequence, including positions that lack sufficient supporting context. Consequently, confidence may favor positions that are not yet ready to be committed.

Recent work identifies a prominent instance of this failure: instruction-tuned DLMs may assign high confidence to end-of-text~(EOT) tokens, causing incomplete or extremely short outputs~\citep{nie2026large}. Existing approaches mitigate this problem through direct EOT suppression~\citep{nie2026large}, additional model training~\citep{kim2025rainbow}, or semi-autoregressive~(semi-AR) decoding~\citep{arriola2025block, cheng2025sdar, wu2025fast}. However, these approaches provide limited gains through token-specific intervention, require additional fine-tuning, or restrict position selection to sequentially generated blocks.

We investigate whether premature EOT generation can instead be mitigated through a minimal inference-time intervention. Inspired by template-infilling methods that insert structures to guide output formats or reasoning patterns~\citep{xiong2025unveiling, jin2025thinking, lee2025unlocking}, we place a short suffix anchor near the end of the response region before decoding. Unlike prior work, which uses inserted anchors to impose detailed output structure, we study whether minimal suffix anchoring can mitigate EOT overconfidence in fully non-AR decoding. We first show that a short anchor, such as \textit{``The answer is''} or even \textit{``.''}, substantially reduces incomplete generation, indicating that detailed structural information is unnecessary for discouraging premature EOT prediction. However, we find that, despite providing no information about the correct response, the anchor induces disproportionately high confidence in nearby positions before sufficient supporting context has been decoded. Confidence-based selection may therefore decode anchor-adjacent tokens too early, such as committing a final answer before the reasoning trajectory is sufficiently developed. We refer to this problem as \textit{anchor-induced local overconfidence}.

To address this problem, we propose \textbf{Suffix-Anchored Confidence Modulation}, a simple training-free method that uses suffix anchoring with progress-dependent confidence modulation near the anchor. Early in decoding, it down-weights confidence scores near the anchor, where anchor-induced overconfidence is strongest, and gradually relaxes this modulation as more context is decoded. This reduces premature commitment of anchor-adjacent tokens while retaining the completion benefit of suffix anchoring.

We evaluate our method across text-only reasoning, vision-language reasoning, and code-generation benchmarks. On LLaDA~\citep{nie2026large} and LLaDA-1.5~\citep{zhu2026llada}, our method consistently improves fully non-AR decoding on GSM8K~\citep{cobbe2021training}, MATH-500~\citep{hendrycks2021measuring, lightman2024let}, StrategyQA~\citep{geva2021did}, and MMLU-Pro~\citep{wang2024mmlu}. On LaViDa~\citep{li2026lavida}, the gains extend to vision-language benchmarks including MathVista~\citep{lu2024mathvista} and ChartQA~\citep{masry2022chartqa}. Our method outperforms diverse baselines covering uniform and confidence-based position selection, remasking, decoding-bias mitigation, direct and training-based EOT mitigation, template infilling, and semi-AR decoding. It also achieves particularly large gains under highly parallel decoding, demonstrating the benefit of improving the fully non-AR decoding over block-wise semi-AR decoding.

% It also provides particularly large gains under limited step budgets, where more tokens must be decoded in parallel, showing that effective fully non-AR decoding can substantially outperform block-wise semi-AR decoding under high parallelism.

\section{Related Work}
\label{sec:related work}

\subsection{Diffusion Language Models}
Diffusion models have achieved strong generative performance in continuous domains such as image and video generation~\citep{sohl2015deep, ho2020denoising, song2020score, karras2022elucidating, peebles2023scalable, ho2022video}. Motivated by this progress, prior work has extended diffusion to discrete text generation through categorical corruption processes, discrete-state Markov chains, and continuous-time variants~\citep{hoogeboom2021argmax, austin2021structured, campbell2022continuous}. Subsequent studies developed masked diffusion language models and clarified connections among different parameterizations~\citep{lou2023discrete, shi2024simplified, sahoo2024simple, zheng2025masked, ou2025your}. Recent large-scale DLMs, including LLaDA~\citep{nie2026large} and LLaDA-1.5~\citep{zhu2026llada}, demonstrate that masked denoising can scale to models with billions of parameters, achieving performance comparable to similar sized autoregressive LLMs while supporting flexible parallel token generation. Building on these advances, our work studies the inference-time position-selection problem in fully non-AR decoding.

\subsection{Decoding Strategies for DLMs}
A standard DLM decoding strategy uses model confidence for position selection, as in top-probability and top-margin decoding~\citep{chang2022maskgit, kim2025train, nie2026large}. Prior work improves DLM decoding through several complementary directions. UNCODE~\citep{huang2026empirical} mitigates decoding bias, while ReMDM~\citep{wang2026remasking} allows previously decoded tokens to be remasked and revised. Template-infilling methods insert predefined output structures or reasoning patterns to guide generation~\citep{xiong2025unveiling, jin2025thinking, lee2025unlocking}, whereas Rainbow Padding~\citep{kim2025rainbow} mitigates the EOT-related failure through additional model training. Semi-AR methods generate blocks sequentially and restrict position selection to the block being generated~\citep{arriola2025block, cheng2025sdar, wu2025fast}, while Fast-dLLM~\citep{wu2025fast} applies confidence-thresholded decoding within this block-wise regime. Our work complements these approaches by studying how suffix anchoring affects confidence-based position selection in fully non-AR decoding. We show that a semantically minimal suffix anchor mitigates premature EOT prediction, identify the resulting anchor-induced local overconfidence, and address it through progress-dependent confidence modulation without additional training or block-wise restrictions.

% A standard training-free strategy for DLM decoding is to use confidence as the position-selection signal, as in top-probability decoding and related variants such as top-margin decoding~\citep{chang2022maskgit, kim2025train, nie2026large}. Recent work also uses confidence for inference-time acceleration and scheduling. Fast-dLLM~\citep{wu2025fast} accelerates DLM inference while limiting quality degradation by applying a confidence threshold and unmasking only positions with sufficiently high prediction confidence. Prophet~\citep{li2025diffusion} observes early answer convergence in DLM trajectories and uses probability gaps between answer candidates to decide when decoding can stop early. ICE~\citep{jin2025thinking} uses in-place chain-of-thought prompting and confidence-aware early exit to improve DLM inference. AdaBlock-dLLM~\citep{lu2025adablock} analyzes confidence dynamics in semi-AR decoding and adaptively adjusts block sizes according to semantic boundary confidence. Together, these works show that confidence is a useful signal for DLM inference. Our work studies a complementary aspect of confidence-based decoding: when and how confidence can mislead position selection in fully non-AR generation, focusing on EOT overconfidence and anchor-induced local overconfidence. We address these issues with a simple training-free modification of standard confidence-based decoding strategies.

\begin{figure*}[t]
\begin{center}
\includegraphics[width=0.955\linewidth]{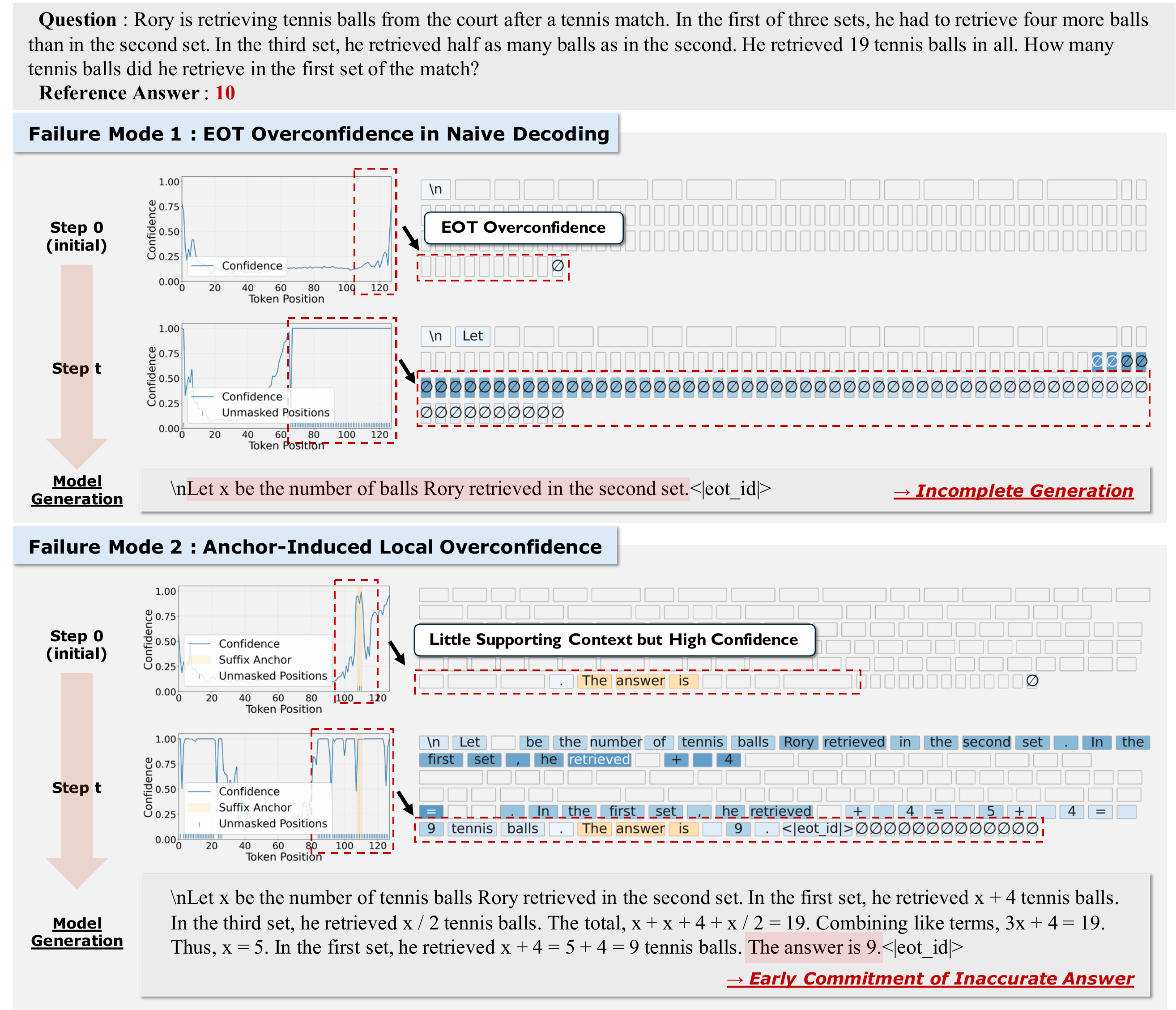}   
    \vspace{-2.0mm}
\end{center}
\caption{\textbf{Two failure modes of confidence-based position selection.} Top: naive confidence-based decoding assigns high confidence to EOT tokens and unmasks them before the response is sufficiently generated, resulting in incomplete output. Bottom: a suffix anchor improves response completion but induces misleadingly high confidence near the anchor despite providing little supporting context, leading to premature decoding of near-anchor tokens. Darker blue token boxes indicate positions decoded at later steps. $\varnothing$ denotes the \texttt{<|endoftext|>} token.}
\label{fig:confidence_failures}
\vspace{-1.3mm}
\end{figure*}
% \caption{\textbf{Two failure modes of confidence-based position selection.} Top: naive confidence-based decoding assigns high confidence to EOT tokens and unmasks them before the response is sufficiently generated, resulting in incomplete output. Bottom: suffix anchoring improves response completion but induces misleadingly high confidence near the anchor, causing anchor-adjacent tokens to be decoded too early and producing an incorrect final answer. Darker blue token boxes indicate positions decoded at later steps. $\varnothing$ denotes the \texttt{<|endoftext|>} token.}

\begin{figure}[]
\begin{center}
\includegraphics[width=0.96\linewidth]{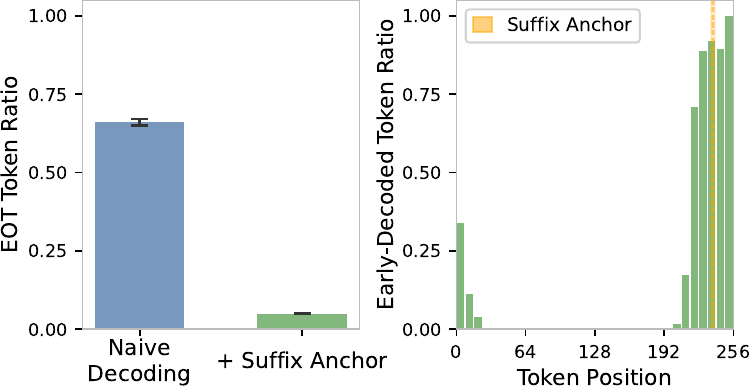}   
    \vspace{-1mm}
\end{center}
\caption{\textbf{Effects of suffix anchoring.} Left: suffix anchoring reduces the EOT token ratio in generated outputs, mitigating incomplete generation. Right: under suffix-anchored decoding, tokens decoded within the first 15\% of steps concentrate near the suffix anchor. The 256-token response region is divided into 32 bins, and each bar reports the average fraction of decoded tokens in the corresponding bin. Yellow vertical lines indicate the suffix-anchor positions. All results are computed on the GSM8K~\citep{cobbe2021training} test split. }
\label{fig:confidence_statistics}
\end{figure}

\section{When Confidence Misleads Position Selection}
\label{sec:preliminary analyses}
\vspace{-1.3mm}

Confidence-based decoding selects masked positions according to their prediction confidence. This decision is particularly important in fully non-AR decoding, where any masked position in the response region can be unmasked, unlike semi-AR decoding, which restricts selection to the current block. In fully non-AR decoding, the candidate set includes distant positions whose supporting context may not yet have been generated, yet some of these positions may nevertheless receive high confidence prematurely. We analyze two instances of this failure. The first is the previously studied problem of EOT overconfidence, which can cause incomplete or extremely short generations~\citep{kim2025rainbow, nie2026large}. 
The second is a failure mode introduced by suffix anchoring: although an anchor discourages premature EOT prediction, it can make nearby positions overconfident despite providing little supporting context. Unless otherwise specified, we use top-probability decoding as the base decoding strategy throughout this analysis.

\paragraph{Failure mode 1: EOT overconfidence in naive decoding.}
In naive fully non-AR decoding, the model may assign high probability to EOT tokens at positions near the end of the response region early in generation. As shown in Figure~\ref{fig:confidence_failures}, confidence-based decoding may therefore unmask these positions before sufficient response content has been generated, resulting in incomplete outputs. This failure has also been reported in recent work~\citep{kim2025rainbow, nie2026large} and demonstrates that high confidence does not necessarily imply that a position should be unmasked.

\paragraph{Suffix anchoring mitigates incomplete generation.}
We examine whether this failure can be mitigated through a minimal inference-time intervention that does not directly suppress EOT tokens. Before decoding, we place a short suffix anchor near the end of the response region, using \textit{``The answer is''} in this analysis. The fixed suffix conditions token predictions throughout the masked response region, encouraging the model to generate non-EOT content up to the anchor and thereby discouraging premature EOT prediction. As shown in Figure~\ref{fig:confidence_statistics}~(left), suffix anchoring substantially reduces the average EOT ratio in generated outputs, indicating fewer incomplete or extremely short generations. Unlike prior template-infilling methods, this short anchor is not intended to impose a detailed output or reasoning structure. Instead, it provides a lightweight suffix conditioning for response continuation. The strong performance of minimal anchors such as \textit{``.''} of \textit{``,''}, reported in Appendix~\ref{app:suffix_anchor_choice}, further supports this interpretation.

\paragraph{Failure mode 2: Anchor-induced local overconfidence.} 
Although suffix anchoring improves response completion, we find that it also distorts confidence near the anchor. As shown in Figure~\ref{fig:confidence_failures}, anchor-adjacent positions can become highly confident before sufficient supporting context has been decoded. Confidence-based decoding therefore tends to unmask these positions too early, producing inaccurate tokens despite their high confidence. Figure~\ref{fig:confidence_statistics}~(right) further shows that a disproportionately large fraction of positions decoded during the first 15\% of the generation process are located near the suffix anchor. This early concentration provides additional evidence that suffix anchoring biases position selection toward the anchor region. We refer to this problem as \textit{anchor-induced local overconfidence}.

\paragraph{Summary.}
We find that even a semantically minimal suffix anchor discourages premature EOT prediction and improves response completion, but also induces early overconfidence near the anchor. This motivates a decoding strategy that retains the completion benefit of suffix anchoring while reducing premature unmasking near the anchor.

\section{Method}
\label{sec:method}

\begin{figure*}[t]
\begin{center}
\includegraphics[width=0.95\linewidth]{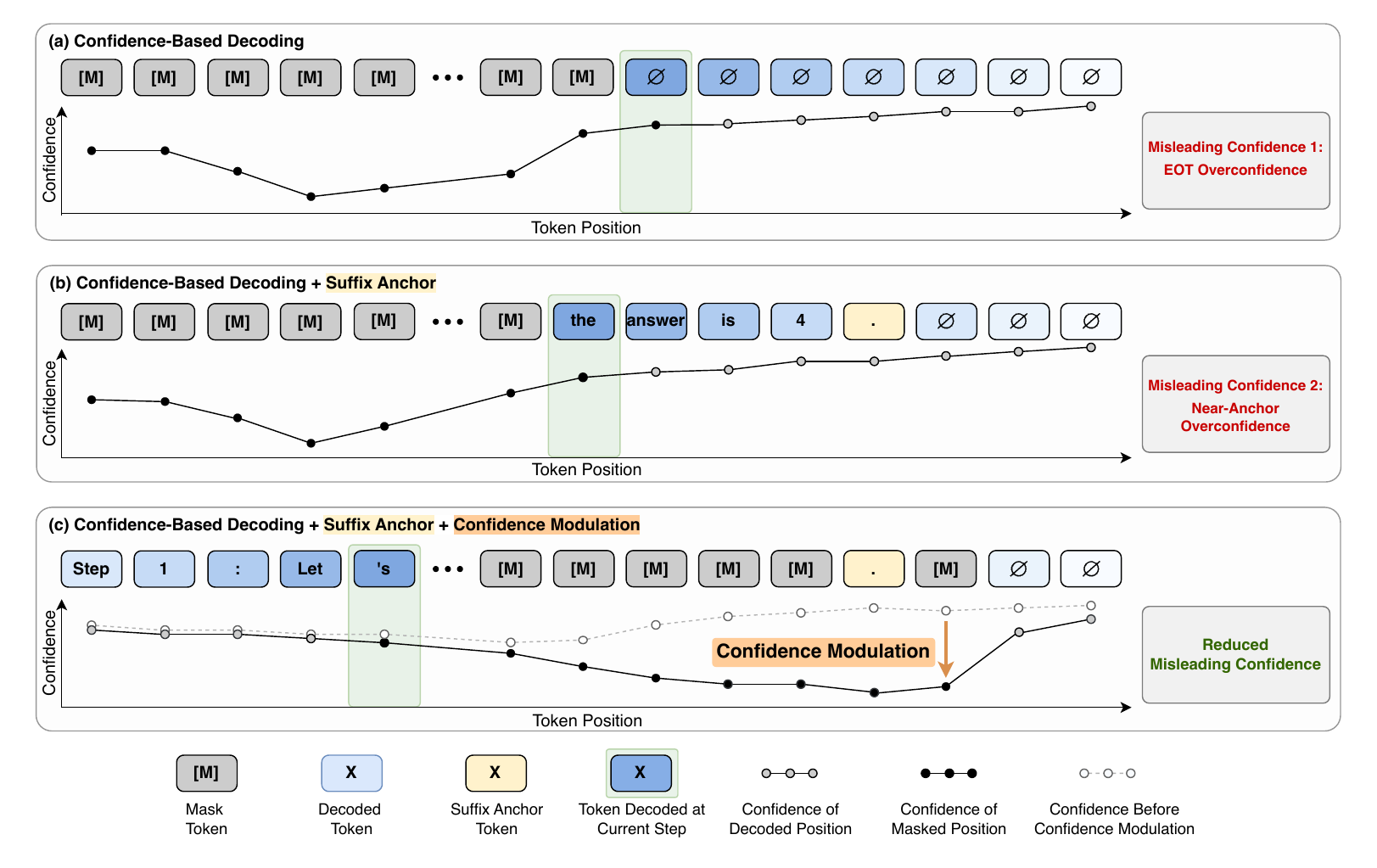}   
    \vspace{-2mm}
\end{center}
\caption{\textbf{Overview of Suffix-Anchored Confidence Modulation.} (a) Standard confidence-based decoding can select high-confidence EOT tokens too early. (b) Adding a suffix anchor reduces EOT overconfidence but may induce misleadingly high confidence near the anchor. (c) Our method applies anchor-proximity confidence modulation to reduce premature decoding of anchor-adjacent positions while preserving the benefit of suffix anchoring. Darker blue token boxes indicate positions decoded at later steps. $\varnothing$ denotes the \texttt{<|endoftext|>} token.}
\label{fig:method}
\vspace{-2mm}
\end{figure*}

We propose Suffix-Anchored Confidence Modulation, a simple training-free decoding method with two components. First, we insert a short suffix anchor to reduce EOT-induced incomplete generation. Second, we down-weight confidence scores near the anchor early in decoding and gradually relax the modulation as decoding progresses, reducing premature decoding of anchor-adjacent positions. Figure~\ref{fig:method} compares standard confidence-based decoding, suffix anchoring alone, and the full method. The complete decoding procedure is outlined in Algorithm~\ref{alg:method} in Appendix~\ref{app:algorithm}.

\paragraph{DLM decoding preliminaries.}
Let $\mathbf{x}^{(t)}=(x_1^{(t)},\ldots,x_L^{(t)})$ be the partially decoded response at decoding step $t$, where $L$ is the response length and unresolved positions are represented by \texttt{[MASK]}. Let $\mathcal{M}^{(t)}=\{i:x_i^{(t)}=\texttt{[MASK]}\}$ be the set of masked positions. At each step, the DLM predicts a token distribution for every masked position,
\begin{equation}
p_\theta(\cdot \mid \mathbf{x}^{(t)}, i), \quad i \in \mathcal{M}^{(t)}.
\end{equation}
A confidence-based decoding strategy assigns each masked position $i$ a confidence score $c_i^{(t)}$, such as the maximum predicted token probability in top-probability decoding or the gap between the top two predicted probabilities in top-margin decoding. The strategy then unmasks a subset of high-confidence positions. Our method reweights $c_i^{(t)}$ after it is computed and can therefore be used with different confidence definitions.

\paragraph{Suffix anchoring.}
Before decoding, we place a short suffix anchor near the end of the response region. The anchor can be a phrase such as \textit{``The answer is’’}, or even a minimal token such as \textit{``.’’} or \textit{``,’’}, as examined in Appendix~\ref{app:suffix_anchor_choice}. Let $\mathcal{A}$ denote the set of anchor positions, which remain fixed throughout decoding. The suffix anchor conditions token predictions throughout the masked response region, encouraging the model to generate non-EOT content up to the anchor and thereby discouraging premature EOT prediction.

\paragraph{Anchor-proximity weight.}
To down-weight confidence near the suffix anchor, we define an anchor-proximity weight for each token position $i$:
\begin{equation}
w_i =
\min\left\{
1,\;
\beta
\max_{a \in \mathcal{A}}
\exp\left(-\frac{|i-a|}{\kappa}\right)
\right\}.
\end{equation}
We use exponential decay as a simple distance-based weighting function that assigns larger weights to positions closer to the anchor, decreases continuously with distance, and remains positive without requiring an explicit cutoff. Here, $\kappa > 0$ controls how rapidly the weight decays with distance from the anchor, while $\beta > 0$ controls the overall modulation strength. Clipping at $1$ keeps the anchor-proximity weight bounded.

\begin{table*}[]
\centering
\resizebox{0.98\textwidth}{!}{%
\begin{tabular}{lll|cc|c|c|c}
\toprule
\multirow{2}{*}{Model}
% & \multirow{2}{*}{\begin{tabular}{@{}c@{}}Decoding\\Regime\end{tabular}}
& \multirow{2}{*}{Decoding Regime}
& \multirow{2}{*}{Decoding Method}
& \multicolumn{2}{c|}{Math Reasoning}
& \multicolumn{1}{c|}{\begin{tabular}{@{}c@{}}Compositional\\Reasoning\end{tabular}}
& \multicolumn{1}{c|}{\begin{tabular}{@{}c@{}}Broad Knowledge\\Reasoning\end{tabular}}
& \multirow{2}{*}{Average} \\
\cmidrule(lr){4-7}
&
&
& \multicolumn{1}{c}{GSM8K}
& \multicolumn{1}{c|}{MATH-500}
& \multicolumn{1}{c|}{StrategyQA}
& \multicolumn{1}{c|}{MMLU-Pro}
& \\
\midrule
\midrule

\multirow{12}{*}{LLaDA}
& \multirow{2}{*}{\begin{tabular}{@{}l@{}}Semi-AR\\(Block Size $=32$)\end{tabular}}
& Semi-AR
& 72.25 & 25.40 & 59.10 & 38.18 & 48.73 \\
&
& Fast-dLLM
& 72.78 & 28.60 & 50.95 & 38.74 & 47.77 \\
\cmidrule(lr){2-8}
&
\multirow{10}{*}{Fully Non-AR}
& Uniform
& 47.16 & 18.00 & 58.81 & 34.26 & 39.56 \\
&
& Top Probability
& 14.94 & 14.60 & 19.65 & 35.23 & 21.11 \\
&
& Top Margin
& 14.78 & 16.20 & 28.53 & 36.39 & 23.98 \\
&
& ReMDM
& 49.96 & 18.40 & 46.00 & 38.39 & 38.19 \\
&
& Direct EOT Suppression
& 52.39 & 20.80 & 64.05 & 37.83 & 43.77 \\
&
& Template Infilling
& 61.94 & 25.20 & 65.21 & 36.74 & 47.27 \\
&
& UNCODE
& 71.34 & 27.60 & 59.68 & 36.57 & 48.80 \\
&
& Rainbow Padding$^{*}$
& 69.37 & 24.60 & 68.56 & 33.93 & 49.12 \\
&
& Ours w/o Confidence Modulation
& 56.18 & 22.40 & 66.08 & 38.30 & 45.74 \\
&
& \textbf{Ours}
& \textbf{76.88} & \textbf{29.00} & \textbf{70.45} & \textbf{39.19} & \textbf{53.88} \\

\midrule

\multirow{12}{*}{LLaDA-1.5}
& \multirow{2}{*}{\begin{tabular}{@{}l@{}}Semi-AR\\(Block Size $=32$)\end{tabular}}
& Semi-AR
& 73.77 & 29.00 & 64.19 & 39.03 & 51.50 \\
&
& Fast-dLLM
& 75.51 & 31.00 & 61.28 & 39.35 & 51.79 \\
\cmidrule(lr){2-8}
&
\multirow{10}{*}{Fully Non-AR}
& Uniform
& 49.81 & 19.40 & 65.36 & 35.70 & 42.57 \\
&
& Top Probability
& 23.35 & 19.40 & 43.67 & 38.54 & 31.24 \\
&
& Top Margin
& 28.58 & 19.80 & 50.95 & 39.05 & 34.60 \\
&
& ReMDM
& 50.49 & 19.00 & 49.20 & 38.58 & 39.32 \\
&
& Direct EOT Suppression
& 51.93 & 19.40 & 66.81 & 38.69 & 44.21 \\
&
& Template Infilling
& 62.02 & 24.60 & 65.21 & 37.45 & 47.32 \\
&
& UNCODE
& 74.00 & 29.40 & 63.17 & 38.06 & 51.16 \\
&
& Rainbow Padding$^{*}$
& 64.14 & 22.60 & 67.98 & 35.59 & 47.58 \\
&
& Ours w/o Confidence Modulation
& 51.40 & 21.00 & 66.67 & 38.55 & 44.41 \\
&
& \textbf{Ours}
& \textbf{75.59} & \textbf{33.40} & \textbf{70.31} & \textbf{39.76} & \textbf{54.77} \\

\bottomrule
\end{tabular}%
}
\caption{\textbf{Results on text-only reasoning benchmarks.}
Accuracy~(\%) is reported on four text-only reasoning benchmarks using
LLaDA 8B-Instruct and LLaDA-1.5.
We compare semi-autoregressive~(Semi-AR) methods with fully non-autoregressive
decoding methods. Ours w/o Confidence Modulation uses suffix anchoring without
confidence modulation, whereas Ours denotes the full method.
\textbf{Bold} indicates the best result for each model on each benchmark.
$^{*}$ indicates a method that requires additional model training; all other
methods are training-free.}
\vspace{-2.8mm}
\label{tab:main_text_results}
\end{table*}

\paragraph{Progress-dependent confidence modulation.}
Anchor-induced overconfidence is most problematic early in decoding, when little supporting context has been decoded. As more tokens are unmasked, anchor-adjacent predictions become conditioned on richer surrounding context. We therefore apply stronger confidence modulation near the suffix anchor early in decoding and gradually relax it over decoding process. Let $m^{(t)}=|\mathcal{M}^{(t)}|$ be the number of masked positions at step $t$. We define decoding progress as
\begin{equation}
p^{(t)} = 1 - \frac{m^{(t)}}{L},
\end{equation}
where larger values indicate later decoding stages. Given the original confidence score $c_i^{(t)}$ from the underlying decoding strategy, we compute the reweighted confidence score as
\begin{equation}
\tilde{c}_i^{(t)}=
c_i^{(t)}
\left(
1 - w_i (1-p^{(t)})^{\gamma}
\right),
\label{eqn:confidence modulation}
\end{equation}
where $\gamma > 0$ controls how quickly the modulation is relaxed. Early in decoding, $(1-p^{(t)})^\gamma$ is large, so confidence near the suffix anchor is down-weighted more strongly. As decoding progresses, this factor decreases toward zero, and $\tilde{c}_i^{(t)}$ approaches the original confidence $c_i^{(t)}$. This reduces premature decoding of anchor-adjacent positions while recovering the base decoding behavior in later stages.

\paragraph{Position selection.}
At each decoding step, the underlying decoding strategy computes a base confidence score $c_i^{(t)}$ for every masked position. We use the reweighted score $\tilde{c}_i^{(t)}$ to rank positions for unmasking, after which the selected positions are filled with the DLM's predicted tokens. Our method requires no model training, auxiliary modules, or architectural changes.
% Because our method only inserts a fixed suffix anchor and reweights scalar confidence scores during decoding, it requires no model training, auxiliary modules, or architectural changes.

\section{Experiments}
\label{sec:experiments}
\vspace{-0.2mm}

\subsection{Experimental Setup}
\label{subsec:experimental setup}

\paragraph{Models and benchmarks.}
We evaluate our method on two representative text-only DLMs, LLaDA 8B-Instruct~\citep{nie2026large} and LLaDA-1.5~\citep{zhu2026llada}, and one vision-language DLM, LaViDa-Instruct~\citep{li2026lavida}. For text-only reasoning, we use GSM8K~\citep{cobbe2021training} and MATH-500~\citep{hendrycks2021measuring, lightman2024let} for mathematical reasoning, StrategyQA~\citep{geva2021did} for commonsense reasoning, and MMLU-Pro~\citep{wang2024mmlu} for broad-domain knowledge reasoning. For vision-language reasoning, we use MathVista~\citep{lu2024mathvista} and ChartQA~\citep{masry2022chartqa}. We additionally evaluate code generation on HumanEval~\citep{chen2021evaluating} and MBPP~\citep{austin2021program}. We use 5-shot prompting for MMLU-Pro, 3-shot prompting for MBPP, and zero-shot prompting for all other benchmarks. We report accuracy for reasoning benchmarks and pass@1 for code-generation benchmarks.

\begin{table}[t]
\centering
\setlength{\tabcolsep}{3.5pt}
\resizebox{\linewidth}{!}{%
\begin{tabular}{lll|cc|c}
\toprule
\multicolumn{1}{l}{Model}
% & \multicolumn{1}{l}{\begin{tabular}{@{}c@{}}Decoding\\Regime\end{tabular}}
& \multicolumn{1}{l}{Decoding Regime}
& \multicolumn{1}{l|}{Decoding Method}
& \multicolumn{1}{c}{MathVista}
& \multicolumn{1}{c|}{ChartQA}
& \multicolumn{1}{c}{Average} \\
\midrule
\midrule

\multirow{11}{*}{LaViDa}
& \multirow{2}{*}{\begin{tabular}{@{}l@{}}Semi-AR\\(Block Size $=32$)\end{tabular}}
& Semi-AR
& 39.50 & 35.88 & 37.69 \\
&
& Fast-dLLM
& 40.30 & 37.28 & 38.79 \\
\cmidrule(lr){2-6}
&
\multirow{9}{*}{Fully Non-AR}
& Uniform
& 39.60 & 27.12 & 33.36 \\
&
& Top Probability
& 40.30 & 24.12 & 32.21 \\
&
& Top Margin
& 39.60 & 23.24 & 31.42 \\
&
& ReMDM
& 42.30 & 29.16 & 35.73 \\
&
& EOT Suppression
& 44.30 & 33.20 & 38.75 \\
&
& Template Infilling
& \textbf{45.60} & 31.12 & 38.36 \\
&
& UNCODE
& 40.80 & 32.64 & 36.72 \\
% &
% & Ours w/o CM
% & 45.50 & 44.96 & 45.23 \\
&
& \textbf{Ours}
& 45.30 & \textbf{46.04} & \textbf{45.67} \\

\bottomrule
\end{tabular}%
}
\caption{\textbf{Results on vision-language reasoning benchmarks.}
Accuracy~(\%) is reported on MathVista and ChartQA using LaViDa-Instruct. \textbf{Bold} indicates the best result on each benchmark.}
\label{tab:main_vl_results}
\vspace{-4mm}
\end{table}

% Ours w/o CM refers to our method without confidence modulation. 

% \begin{table}[t]
% \centering
% \setlength{\tabcolsep}{4pt}
% \resizebox{\linewidth}{!}{%
% \begin{tabular}{ll|cc|c}
% \toprule
% \multicolumn{1}{l}{Model}
% & \multicolumn{1}{l|}{Decoding Method}
% & \multicolumn{1}{c}{MathVista}
% & \multicolumn{1}{c|}{ChartQA}
% & \multicolumn{1}{c}{Average} \\
% \midrule
% \midrule
% \multirow{7}{*}{LaViDa}
% & Random                         & 28.80 & 27.12 & 27.96 \\
% \cmidrule(lr){2-5}
% & Top Probability                & 27.00 & 24.12 & 25.56 \\
% & \quad +  Suffix Anchor  & 29.10 & 44.96 & 37.03 \\
% & \quad + Confidence Modulation   & \textbf{34.60} & \textbf{45.92} & \textbf{40.26} \\
% \cmidrule(lr){2-5}
% & Top Margin                     & 24.80 & 23.24 & 24.02 \\
% & \quad +  Suffix Anchor  & 29.20 & 45.08 & 37.14 \\
% & \quad + Confidence Modulation   & \textbf{33.20} & \textbf{45.44} & \textbf{39.32} \\
% \bottomrule
% \end{tabular}%
% }
% \caption{\textbf{Results on vision-language reasoning benchmarks.} Accuracy~(\%) is reported on MathVista and ChartQA using LaViDa-Instruct. For each confidence-based decoding strategy, the unmodified baseline, suffix anchoring, and the full method with confidence modulation are compared. \textbf{Bold} indicates the best result within each confidence-based decoding group.}
% \label{tab:main_vl_results}
% \vspace{-4mm}
% \end{table}
% Random position selection is included as a non-confidence-based reference.

\paragraph{Baselines.}
We compare against semi-AR and fully non-AR decoding baselines. For semi-AR decoding, we evaluate the default top-probability~\citep{chang2022maskgit, nie2026large} strategy and Fast-dLLM~\citep{wu2025fast} with a block size of 32. For fully non-AR decoding, we include Uniform, Top Probability, Top Margin~\citep{kim2025train}, ReMDM~\citep{wang2026remasking}, Direct EOT Suppression~\citep{nie2026large}, Template Infilling, UNCODE~\citep{huang2026empirical}, and Rainbow Padding~\citep{kim2025rainbow}. Uniform selects positions randomly, while Direct EOT Suppression assigns negative-infinity confidence to EOT predictions. For Template Infilling, we follow prior work~\citep{jin2025thinking} and use the template \textit{``Step 1: \ldots Step 2: \ldots Step 3: \ldots The answer is: \ldots''}. We use the reported hyperparameters for UNCODE, ReMDM, and Fast-dLLM, and train Rainbow Padding using its official implementation.

\paragraph{Implementation details.}
We use a response length $L=256$ for GSM8K and MATH-500, and $L=128$ for the remaining benchmarks. Unless otherwise specified, the number of decoding steps is set to $T=L/2$, except for Fast-dLLM, which instead uses a confidence threshold of $\tau=0.9$ for unmasking. The suffix anchor is placed 20 positions before the end of the response region, leaving most of the response budget available for free-form generation. In the main experiments, we use \textit{``The answer is''} as the suffix anchor for reasoning tasks and \texttt{return} for code generation. Our method is applied to top-probability decoding unless otherwise specified, while top-margin results are reported in Appendix~\ref{app:top_margin_results}. Following standard DLM evaluation protocols~\citep{nie2026large, li2025diffusion, kim2025rainbow}, all methods use deterministic argmax decoding with temperature 0. Additional implementation details are provided in Appendix~\ref{app:experimental_details}.

\paragraph{Hyperparameters.}
Our method uses three hyperparameters: $\kappa$, $\beta$, and $\gamma$. We select them with a lightweight sweep over $\kappa \in \{12,14\}$, $\beta \in \{1.0,1.1,1.2,1.3,1.4,1.5\}$, and $\gamma \in \{0.7,0.85,1.0\}$ on a small subset of 128 samples from the training or validation split when available. The sweep is conducted with LLaDA under top-probability decoding, and the selected values are reused for LLaDA-1.5 on the same benchmark. When no training or validation split is available, we use the GSM8K setting, $(\kappa,\beta,\gamma)=(14,1.3,0.85)$. Details, including sensitivity analysis, are provided in Appendix~\ref{app:hyperparameter}.

\subsection{Main Results}
\label{subsec:main results and analysis}

\begin{table}[t]
\centering
\setlength{\tabcolsep}{4pt}
\resizebox{\linewidth}{!}{%
\begin{tabular}{ll|cccc}
\toprule
\multirow{2}{*}{\begin{tabular}{@{}l@{}}Decoding\\Regime\end{tabular}}
& \multirow{2}{*}{Method / Configuration}
& \multicolumn{4}{c}{Step Budget~($T$)} \\
\cmidrule(lr){3-6}
&
& $32$ & $64$ & $128$ & $256$ \\
\midrule

\multirow{4}{*}{Fully Non-AR}
& Uniform
& 39.04 & 44.50 & 47.16 & 51.48 \\
& Top Probability
& 32.52 & 26.46 & 14.94 & 2.96 \\
& Top Margin
& 34.04 & 25.09 & 14.78 & 7.35 \\
& \textbf{Ours}
& \textbf{57.70} & \textbf{69.67} & \textbf{76.88} & \textbf{77.41} \\

\midrule

\multirow{4}{*}{Semi-AR}
& Block Size $=8$
& 11.37 & 56.71 & 72.33 & \underline{74.07} \\
& Block Size $=16$
& 21.15 & 60.96 & \underline{73.39} & \underline{74.07} \\
& Block Size $=32$
& 29.72 & 63.31 & 72.25 & 73.01 \\
& Block Size $=64$
& \underline{36.32} & \underline{63.46} & 72.18 & 71.87 \\

\bottomrule
\end{tabular}%
}
\caption{\textbf{Parallel decoding capability of fully non-AR and semi-AR decoding.}
GSM8K accuracy~(\%) is reported using LLaDA with response length $L=256$. All methods unmask $L/T$ tokens per step, so smaller $T$ indicates higher parallelism. Our fully non-AR method outperforms all semi-AR configurations across step budgets, with the largest margin at the most highly parallel setting, $T=32$. \textbf{Bold} and \underline{underlining} indicate the best overall and semi-AR results, respectively.}
\vspace{-1mm}
\label{tab:semi_ar_comparison}
\end{table}

\paragraph{Text-only reasoning.}
Table~\ref{tab:main_text_results} reports results on four text-only reasoning benchmarks using LLaDA and LLaDA-1.5. Our method achieves the best average accuracy for both models, reaching $53.88$ on LLaDA and $54.77$ on LLaDA-1.5. These results exceed the strongest competing baselines, Rainbow Padding and Fast-dLLM, by $4.76$ and $2.98$ points, respectively. Notably, these baselines require additional training or block-wise semi-AR decoding, whereas our method is training-free and fully non-AR. Our method also achieves the best result on every benchmark for both models. The comparison with Ours w/o Confidence Modulation further shows that suffix anchoring provides substantial gains and confidence modulation consistently improves upon it. Results with top-margin decoding are reported in Appendix~\ref{app:top_margin_results}.

\paragraph{Vision-language and code generation.}
Table~\ref{tab:main_vl_results} reports results on MathVista and ChartQA using LaViDa. Our method achieves the best average accuracy of $45.67$, outperforming the strongest competing baseline, Fast-dLLM, by $6.88$ points. It obtains the best result on ChartQA with $46.04$ accuracy and remains competitive on MathVista, achieving $45.30$ compared with $45.60$ for Template Infilling. Code-generation results on HumanEval and MBPP are provided in Table~\ref{tab:code_generation_results} of Appendix~\ref{app:code_generation}.

\paragraph{Fully non-AR advantage under high parallelism.}
Table~\ref{tab:semi_ar_comparison} compares our method with semi-AR decoding across different step budgets and block sizes. Since all methods unmask $L/T$ tokens per step, smaller $T$ requires more tokens to be decoded in parallel. Our method outperforms every semi-AR configuration across all step budgets, with the largest margin at $T=32$, the most highly parallel setting. At this step budget, the best semi-AR configuration achieves $36.32$, while performance drops to $11.37$ as the block size decreases from 64 to 8. In contrast, our method reaches $57.70$ by allowing position selection across the entire response region. These results suggest that restricting selection to short semi-AR response blocks becomes particularly limiting under high parallelism, whereas sequence-wide selection in the fully non-AR regime better supports parallel decoding.

\subsection{Ablation Studies and Efficiency Analysis}
\label{subsec:ablation studies and efficiency analysis}
Unless otherwise specified, we conduct ablation studies and efficiency analysis on GSM8K using LLaDA 8B-Instruct with top-probability decoding. 

\begin{table}[]
\centering
\setlength{\tabcolsep}{7pt}
\resizebox{0.74\linewidth}{!}{%
\begin{tabular}{l|c}
\toprule
Method & Accuracy \\
\midrule
Ours w/o Progress Dependence & 72.25 \\
Ours & \textbf{76.88} \\
\bottomrule
\end{tabular}%
}
\vspace{-0.9mm}
\caption{\textbf{Ablation of progress dependence in confidence modulation.}}
\vspace{-1.0mm}
\label{tab:progress_ablation}
\end{table}

\begin{table}[]
\centering
\setlength{\tabcolsep}{13pt}
\resizebox{0.88\linewidth}{!}{%
\begin{tabular}{l|ccc}
\toprule
\multirow{2}{*}{Decoding Method}
& \multicolumn{3}{c}{Response Length~(L)} \\
\cmidrule(lr){2-4}
& $64$ & $128$ & $256$ \\
\midrule
Top Probability & 39.35 & 23.58 & 14.94 \\
% Ours w/o Confidence Modulation & 51.55 & 56.94 & 49.89 \\
Ours & \textbf{62.55} & \textbf{75.82} & \textbf{76.88} \\
\bottomrule
\end{tabular}
}
\vspace{-0.9mm}
\caption{\textbf{Ablation over response length.} Response length is varied over $L \in \{64,128,256\}$, with the decoding step budget set to $T=L/2$.}
\vspace{-2mm}
\label{tab:length_ablation}
\end{table}

\begin{table}[h]
\centering
\setlength{\tabcolsep}{9pt}
\resizebox{\linewidth}{!}{%
\begin{tabular}{l|l|l|c}
\toprule
Suffix Anchor & Model & Decoding Method & Accuracy \\
\midrule
\midrule

\multirow{2}{*}{No Suffix Anchor}
& LLaDA     & Top Probability & 14.94 \\
\cmidrule(lr){2-4}
& LLaDA-1.5 & Top Probability & 23.35 \\

\midrule

\multirow{4}{*}{\begin{tabular}{@{}c@{}}\textit{``The answer is''}\\(Default)\end{tabular}}
& \multirow{2}{*}{LLaDA}
& Ours w/o CM & 49.89 \\
&
& Ours                         & \textbf{76.88} \\
\cmidrule(lr){2-4}
& \multirow{2}{*}{LLaDA-1.5}
& Ours w/o CM & 51.40 \\
&
& Ours                         & \textbf{75.59} \\

\midrule

% \multirow{4}{*}{\textit{``,''}}
% & \multirow{2}{*}{LLaDA}
% & Ours w/o CM & 55.12 \\
% &
% & Ours                         & \textbf{73.77} \\
% \cmidrule(lr){2-4}
% & \multirow{2}{*}{LLaDA-1.5}
% & Ours w/o CM & 51.86 \\
% &
% & Ours                         & \textbf{72.18} \\

% \midrule

\multirow{4}{*}{\textit{``.''}}
& \multirow{2}{*}{LLaDA}
& Ours w/o CM & 54.13 \\
&
& Ours                         & \textbf{74.68} \\
\cmidrule(lr){2-4}
& \multirow{2}{*}{LLaDA-1.5}
& Ours w/o CM & 53.53 \\
&
& Ours                         & \textbf{75.36} \\

\bottomrule
\end{tabular}%
}
\vspace{-0.8mm}
\caption{\textbf{Ablation over suffix anchors.}
Accuracy~(\%) is reported on GSM8K using LLaDA 8B-Instruct and LLaDA-1.5 with top-probability decoding. Each suffix anchor is inserted at the same anchor position before decoding begins. Ours w/o CM refers to our method without confidence modulation. \textbf{Bold} indicates the better result within each model and suffix-anchor setting. The full table is provided in Table~\ref{tab:suffix_anchor_ablation} of Appendix~\ref{app:suffix_anchor_choice}}
\label{tab:short_suffix_anchor_ablation}
\vspace{-1.7mm}
\end{table}
\begin{table}[]
\centering
\setlength{\tabcolsep}{12.5pt}
\resizebox{0.93\linewidth}{!}{%
\begin{tabular}{l|cc}
\toprule
Decoding Method
& \begin{tabular}{@{}c@{}}Throughput~$\uparrow$\end{tabular}
& \begin{tabular}{@{}c@{}}Latency~$\downarrow$\end{tabular} \\
\midrule
Top Probability & 25.02 & 10.23 \\
Ours & 24.93 & 10.27 \\
\bottomrule
\end{tabular}%
}
\vspace{-0.5mm}
\caption{\textbf{Inference efficiency.} Throughput~(tokens/s) is the average number of generated tokens per second, and latency~(s/sample) is the average inference time per sample. All measurements are taken on a single NVIDIA A6000 GPU.}
\vspace{-2.5mm}
\label{tab:efficiency}
\end{table}

% Ours w/o Confidence Modulation & 25.03 & 10.23 \\

\paragraph{Progress-dependent modulation.}
Table~\ref{tab:progress_ablation} evaluates the effect of the progress-dependent factor $(1-p^{(t)})^\gamma$ in Eq.~(\ref{eqn:confidence modulation}). Removing this factor keeps the confidence modulation fixed throughout decoding and reduces accuracy from $76.88$ to $72.25$. This result supports that gradually relaxing the modulation as decoding progresses is beneficial. 

\paragraph{Response length.}
Table~\ref{tab:length_ablation} reports results across different response lengths with the step budget set to $T=L/2$. Our method achieves strong performance across all settings, indicating robustness to the response length.

\paragraph{Suffix-anchor choice.}
Table~\ref{tab:short_suffix_anchor_ablation} compares the default anchor with the semantically minimal anchor, \textit{``.''} using LLaDA and LLaDA-1.5. With the full method, the minimal anchor performs close to \textit{``The answer is''}, indicating that the improvement does not require substantial semantic content in the anchor. Complete ablations of anchor choices and positions are provided in Appendix~\ref{app:suffix_anchor_choice} and Appendix~\ref{app:anchor_position}, respectively.

\paragraph{Inference efficiency.}
Table~\ref{tab:efficiency} shows that our method introduces negligible inference overhead. Compared with top-probability decoding, throughput changes from $25.02$ to $24.93$ tokens/s and latency from $10.23$ to $10.27$ s/sample. This shows that our method improves decoding quality while largely preserving inference efficiency.

% We conduct ablation studies and efficiency analysis on GSM8K using LLaDA 8B-Instruct with top-probability decoding.

% \paragraph{Effect of progress dependence in confidence modulation.}
% Table~\ref{tab:progress_ablation} evaluates the effect of the progress-dependent factor $(1-p^{(t)})^\gamma$ in Eq.~(\ref{eqn:confidence modulation}). Without this factor, the anchor-proximity confidence down-weighting remains fixed throughout decoding, reducing accuracy from $76.88$ to $72.25$. This shows that gradually relaxing the confidence down-weighting as decoding progresses helps improve performance.

% \paragraph{Effect of generation length.}
% Table~\ref{tab:length_ablation} reports results across different generation lengths, with the step budget set to $T=L/2$. Suffix anchoring improves over the unmodified baseline across all generation lengths, and the full method with confidence modulation further improves performance in every setting.

% \paragraph{Inference efficiency.}
% Table~\ref{tab:efficiency} reports inference throughput and latency. Suffix anchoring and confidence modulation introduce negligible overhead compared with the baseline: throughput remains around $24.8$ tokens/sec and latency remains around $10.3$ seconds per sample for all three decoding variants. Thus, our method improves decoding quality without sacrificing inference efficiency.

\section{Conclusion}
\label{sec:conclusion}
In this work, we studied how confidence-based position selection can fail in fully non-AR DLM decoding through premature EOT prediction and anchor-induced local overconfidence. We proposed Suffix-Anchored Confidence Modulation, which uses suffix anchoring together with progress-dependent confidence modulation near the anchor. Across text-only reasoning, vision-language reasoning, and code-generation benchmarks, our method consistently outperforms diverse semi-AR and fully non-AR baselines. It also remains effective under highly parallel decoding, highlighting the benefit of improving sequence-wide position selection in the fully non-AR regime.
\section*{Limitations}
Our method modifies confidence-based position selection without updating model parameters. It therefore does not address errors caused by insufficient model knowledge or reasoning ability and may provide smaller gains when failures arise primarily from incorrect token predictions rather than premature or suboptimal position selection.

For simplicity, the main experiments use fixed suffix anchors and a predefined anchor position. However, Table~\ref{tab:suffix_anchor_ablation} in Appendix~\ref{app:suffix_anchor_choice} shows that our method remains robust across different suffix anchor choices, and Table~\ref{tab:anchor_position_ablation} in Appendix~\ref{app:anchor_position} shows that it also remains robust across different anchor positions within the later response region. While these two ablations indicate robustness to anchor form and placement, the optimal anchor form or placement may still vary across tasks and output formats. In addition, confidence modulation introduces a small number of hyperparameters, which we tune with a lightweight sweep and reuse across settings when possible. More adaptive strategies for automatically choosing anchors, anchor placements, and modulation strengths would be a valuable direction for future work.

Finally, our experiments focus on representative text-only and vision-language DLMs using standard reasoning and code-generation benchmarks. Further evaluation on multilingual tasks and more diverse multimodal settings would be valuable.

% Comment the below line for under review
\section*{Acknowledgments}

This work was supported by Institute of Information \& communications Technology Planning \& Evaluation (IITP) grant funded by the Korea government (MSIT) ([NO.RS-2021-II211343, Artificial Intelligence Graduate School Program (Seoul National University)], [No.RS-2023-00235293, Development of autonomous driving big data processing, management, search, and sharing interface technology to provide autonomous driving data according to the purpose of usage]) and the InnoCORE program of the Ministry of Science and ICT (26-InnoCORE-01).

\bibliography{bibliography}

\clearpage
\appendix

\addtocontents{toc}{\protect\setcounter{tocdepth}{2}}
\tableofcontents
\clearpage
\section{Algorithm}
\label{app:algorithm}

Algorithm~\ref{alg:method} summarizes the complete decoding procedure for Suffix-Anchored Confidence Modulation. Starting from a masked response sequence, our method first inserts a suffix anchor at predefined response positions and computes the anchor-proximity weights. At each decoding step, the underlying confidence-based strategy computes token predictions and confidence scores for the remaining masked positions. Our method then reweights these confidence scores according to anchor proximity and decoding progress, while leaving the base position-selection rule and token prediction rule unchanged.

\begin{algorithm*}[]
    \caption{Suffix-Anchored Confidence Modulation}
    \label{alg:method}
    \begin{algorithmic}[1]
        \REQUIRE Model $M_\theta$, prompt $\mathbf{x}_{\mathrm{prompt}}$, response length $L$, decoding step budget $T$
        \REQUIRE Suffix anchor tokens $\mathbf{x}_{\mathrm{anchor}}$, anchor positions $\mathcal{A}$
        \REQUIRE Base confidence function $C(\cdot)$, position-selection rule $\mathrm{Select}(\cdot)$
        \REQUIRE Hyperparameters $\kappa$, $\beta$, $\gamma$
        \STATE Initialize $\mathbf{x}^{(T)} \leftarrow \mathrm{InsertAnchor}(\mathrm{concat}(\mathbf{x}_{\mathrm{prompt}}, [\mathrm{MASK}]^L), \mathbf{x}_{\mathrm{anchor}}, \mathcal{A})$
        \STATE Compute anchor-proximity weights for all response positions $i$:
        \STATE \hspace{1em} $w_i \leftarrow \min\left\{1,\; \beta \max\limits_{a \in \mathcal{A}} \exp\left(-\frac{|i-a|}{\kappa}\right)\right\}$
        \FOR{$t = T, T-1, \ldots, 1$}
            \STATE $\mathcal{M}^{(t)} \leftarrow \{i: x_i^{(t)} = [\mathrm{MASK}]\}$
            \STATE Compute logits $\mathbf{z}^{(t)} \leftarrow M_\theta(\mathbf{x}^{(t)})$
            \STATE Predict tokens $\hat{\mathbf{x}}_0 \leftarrow \arg\max(\mathbf{z}^{(t)}, \mathrm{dim}=-1)$
            \STATE Compute base confidence scores $c_i^{(t)} \leftarrow C(\mathbf{z}^{(t)}, i)$ for all $i \in \mathcal{M}^{(t)}$
            \STATE Compute decoding progress $p^{(t)} \leftarrow 1 - |\mathcal{M}^{(t)}|/L$
            \STATE Reweight confidence scores:
            \STATE \hspace{1em} $\tilde{c}_i^{(t)} \leftarrow c_i^{(t)} \left(1 - w_i(1-p^{(t)})^\gamma\right)$ for all $i \in \mathcal{M}^{(t)}$
            \STATE Select positions to unmask $\mathcal{U}^{(t)} \leftarrow \mathrm{Select}(\{\tilde{c}_i^{(t)}\}_{i \in \mathcal{M}^{(t)}})$
            \STATE Update $\mathbf{x}^{(t-1)} \leftarrow \mathbf{x}^{(t)}$
            \STATE Replace $x_i^{(t-1)} \leftarrow \hat{x}_{0,i}$ for all $i \in \mathcal{U}^{(t)}$
        \ENDFOR
        \STATE \textbf{Return:} response segment of $\mathbf{x}^{(0)}$
    \end{algorithmic}
\end{algorithm*}

\section{Additional Experimental Details}
\label{app:experimental_details}

\subsection{Models and Evaluation Splits}
\label{app:models_and_splits}

\begin{table}[h]
\centering
\setlength{\tabcolsep}{5pt}
\resizebox{\linewidth}{!}{%
\begin{tabular}{l|l|c|c}
\toprule
Benchmark & Hugging Face Identifier & Split & Size \\
\midrule
GSM8K & openai/gsm8k & test & 1,319 \\
MATH-500 & HuggingFaceH4/MATH-500 & test & 500 \\
StrategyQA & ChilleD/StrategyQA & test & 687 \\
MMLU-Pro & TIGER-Lab/MMLU-Pro & test & 12,032 \\
MathVista & AI4Math/MathVista & testmini & 1,000 \\
ChartQA & HuggingFaceM4/ChartQA & test & 2,500 \\
HumanEval & openai/openai\_humaneval & test & 164 \\
MBPP & google-research-datasets/mbpp & test & 500 \\
\bottomrule
\end{tabular}%
}
\caption{\textbf{Evaluation datasets and splits.} Hugging Face identifiers, evaluation splits, and evaluation-set sizes used in our experiments. MathVista uses the testmini split because answer labels are not available for the test split.}
\label{tab:datasets_splits}
\end{table}

We use the publicly available checkpoints GSAI-ML/LLaDA-8B-Instruct, GSAI-ML/LLaDA-1.5, and jacklishufan/lavida-llada-v1.0-instruct for LLaDA 8B-Instruct, LLaDA-1.5, and LaViDa-Instruct, respectively. Table~\ref{tab:datasets_splits} summarizes the datasets, Hugging Face identifiers, evaluation splits, and evaluation-set sizes used in our experiments. We use the test split for all benchmarks except MathVista, for which we use the testmini split because answer labels are not provided for the test split.

\subsection{Prompting and Evaluation Protocol}
\label{app:prompting_protocol}
We implement DLM evaluation on the reported benchmarks based on the evaluation setup of lm-evaluation-harness~\citep{eval-harness}, simple-evals~\citep{simple-evals}, and the LaViDa~\citep{li2026lavida} codebase. For multiple-choice benchmarks, we use generative evaluation: the model generates a response, and the final answer is extracted from the generated text rather than selecting among answer candidates by log probability. For reasoning benchmarks, we include \textit{``Let's think step by step.''} or an equivalent task-specific instruction to elicit intermediate reasoning before the final answer. For code-generation benchmarks, we follow the corresponding benchmark-specific prompting and execution-based evaluation protocols. Table~\ref{tab:prompt_template} provides the prompt template used for each benchmark.

\subsection{Hyperparameter Selection and Sensitivity Analysis}
\label{app:hyperparameter}
\begin{table}[h]
\centering
\setlength{\tabcolsep}{12pt}
\resizebox{0.75\linewidth}{!}{%
\begin{tabular}{l|ccc}
\toprule
Benchmark & $\kappa$ & $\beta$ & $\gamma$ \\
\midrule
GSM8K & 14 & 1.3 & 0.85 \\
MATH-500 & 12 & 1.5 & 0.85 \\
StrategyQA & 12 & 1.3 & 0.70 \\
MMLU-Pro & 14 & 1.3 & 0.85 \\
MathVista & 14 & 1.3 & 0.85 \\
ChartQA & 12 & 1.0 & 1.00 \\
HumanEval & 14 & 1.3 & 0.85 \\
MBPP & 14 & 1.5 & 0.70 \\
\bottomrule
\end{tabular}%
}
\caption{\textbf{Selected hyperparameters.} Hyperparameter values selected for each benchmark. The GSM8K setting is used for benchmarks without a training or validation split.}  
\label{tab:selected_hyperparameters}
\end{table}
When a training or validation split is available, we select the hyperparameters $(\kappa,\beta,\gamma)$ using a lightweight sweep over 128 randomly sampled examples. The sweep range is $\kappa \in \{12,14\}$, $\beta \in \{1.0,1.1,1.2,1.3,1.4,1.5\}$, and $\gamma \in \{0.7,0.85,1.0\}$. The sweep is conducted with LLaDA 8B-Instruct under top-probability decoding, and the selected values are reused for top-margin decoding and Dream experiments on the same benchmark. For benchmarks without a training or validation split, we use the GSM8K setting. Table~\ref{tab:selected_hyperparameters} reports the selected hyperparameters for each benchmark.

\begin{table}[h]
\centering
\setlength{\tabcolsep}{10pt}
\resizebox{0.82\linewidth}{!}{%
\begin{tabular}{c|ccc|c}
\toprule
Varied & $\kappa$ & $\beta$ & $\gamma$ & Accuracy \\
\midrule
\multirow{6}{*}{$\kappa$}
& 6 & 1.3 & 0.85 & 73.83 \\
& 8 & 1.3 & 0.85 & 78.52 \\
& 10 & 1.3 & 0.85 & 76.96 \\
& 12 & 1.3 & 0.85 & 80.08 \\
& \textbf{14} & \textbf{1.3} & \textbf{0.85} & \textbf{82.03} \\
& 16 & 1.3 & 0.85 & 79.30 \\
\midrule
\multirow{6}{*}{$\beta$}
& 14 & 1.0 & 0.85 & 78.13 \\
& 14 & 1.1 & 0.85 & 77.74 \\
& 14 & 1.2 & 0.85 & 80.08 \\
& \textbf{14} & \textbf{1.3} & \textbf{0.85} & \textbf{82.03} \\
& 14 & 1.4 & 0.85 & 79.30 \\
& 14 & 1.5 & 0.85 & 80.47 \\
\midrule
\multirow{6}{*}{$\gamma$}
& 14 & 1.3 & 0.40 & 79.30 \\
& 14 & 1.3 & 0.55 & 80.47 \\
& 14 & 1.3 & 0.70 & 81.25 \\
& \textbf{14} & \textbf{1.3} & \textbf{0.85} & \textbf{82.03} \\
& 14 & 1.3 & 1.00 & 80.86 \\
& 14 & 1.3 & 1.15 & 78.52 \\
\bottomrule
\end{tabular}%
}
\caption{\textbf{Hyperparameter sensitivity on GSM8K.} Accuracy is measured on 256 randomly sampled training examples using LLaDA 8B-Instruct with top-probability decoding. Each hyperparameter is varied around the GSM8K setting $(\kappa,\beta,\gamma)=(14,1.3,0.85)$.}
\label{tab:hyperparameter_sensitivity}
\end{table}

To assess sensitivity, we vary each hyperparameter around the GSM8K setting on a randomly sampled subset of 256 training examples, using LLaDA 8B-Instruct with top-probability decoding. For $\kappa$ and $\gamma$, we additionally include values outside the selection sweep to test robustness to wider ranges. Table~\ref{tab:hyperparameter_sensitivity} shows that performance remains stable across a wide range of values. On the same subset, unmodified top-probability decoding and suffix anchoring alone obtain $15.63$ and $55.86$, respectively; all hyperparameter settings in Table~\ref{tab:hyperparameter_sensitivity} substantially exceed these scores. This suggests that the gains of our method do not rely on a narrowly tuned hyperparameter choice.

\section{Additional Experiments}

\subsection{Applicability to Top-Margin Decoding}
\label{app:top_margin_results}

\begin{table*}[]
\centering
\resizebox{0.98\textwidth}{!}{
\begin{tabular}{lll|cc|c|c|c}
\toprule
\multirow{2}{*}{Model}
& \multirow{2}{*}{\begin{tabular}{@{}l@{}}Base Decoding\\Strategy\end{tabular}}
& \multirow{2}{*}{Decoding Method}
& \multicolumn{2}{c|}{Math Reasoning}
& \multicolumn{1}{c|}{\begin{tabular}{@{}c@{}}Compositional\\Reasoning\end{tabular}}
& \multicolumn{1}{c|}{\begin{tabular}{@{}c@{}}Broad Knowledge\\Reasoning\end{tabular}}
& \multirow{2}{*}{Average} \\
\cmidrule(lr){4-7}
&
&
& \multicolumn{1}{c}{GSM8K}
& \multicolumn{1}{c|}{MATH-500}
& \multicolumn{1}{c|}{StrategyQA}
& \multicolumn{1}{c|}{MMLU-Pro}
& \\
\midrule
\midrule

\multirow{6}{*}{LLaDA}
& \multirow{3}{*}{\begin{tabular}{@{}l@{}}Top Probability\\ (Default)\end{tabular}}
& Baseline
& 14.94 & 14.60 & 19.65 & 35.23 & 21.11 \\
&
& Ours w/o Confidence Modulation
& 56.18 & 22.40 & 66.08 & 38.30 & 45.74 \\
&
& \textbf{Ours}
& \textbf{76.88} & \textbf{29.00} & \textbf{70.45} & \textbf{39.19} & \textbf{53.88} \\
\cmidrule(lr){2-8}
&
\multirow{3}{*}{Top Margin}
& Baseline
& 14.78 & 16.20 & 28.53 & 36.39 & 23.98 \\
&
& Ours w/o Confidence Modulation
& 56.18 & 22.40 & 66.08 & 38.30 & 45.74 \\
&
& \textbf{Ours}
& \textbf{72.33} & \textbf{25.40} & \textbf{68.12} & \textbf{38.42} & \textbf{51.07} \\

\midrule

\multirow{6}{*}{LLaDA-1.5}
& \multirow{3}{*}{\begin{tabular}{@{}l@{}}Top Probability\\(Default)\end{tabular}}
& Baseline
& 23.35 & 19.40 & 43.67 & 38.54 & 31.24 \\
&
& Ours w/o Confidence Modulation
& 51.40 & 21.00 & 66.67 & 38.55 & 44.41 \\
&
& \textbf{Ours}
& \textbf{75.59} & \textbf{33.40} & \textbf{70.31} & \textbf{39.76} & \textbf{54.77} \\
\cmidrule(lr){2-8}
&
\multirow{3}{*}{Top Margin}
& Baseline
& 28.58 & 19.80 & 50.95 & 39.05 & 34.60 \\
&
& Ours w/o Confidence Modulation
& 54.66 & 21.20 & 66.08 & 38.72 & 45.17 \\
&
& \textbf{Ours}
& \textbf{70.96} & \textbf{25.80} & \textbf{68.56} & \textbf{39.39} & \textbf{51.18} \\

\bottomrule
\end{tabular}
}
\caption{\textbf{Applicability to different confidence-based decoding strategies.}
Accuracy~(\%) is reported on four text-only reasoning benchmarks using
LLaDA 8B-Instruct and LLaDA-1.5.
We apply suffix anchoring and confidence modulation on top of either
top-probability or top-margin decoding.
The baseline uses the corresponding base decoding strategy without suffix
anchoring or confidence modulation.
Ours w/o Confidence Modulation uses suffix anchoring alone, whereas Ours
denotes the full method.
The results show that the proposed method consistently improves both base
decoding strategies across the two models.
\textbf{Bold} indicates the best result within each model and base decoding
strategy.}
\vspace{-2.8mm}
\label{tab:confidence_strategy_results}
\end{table*}

In the main experiments, we apply our method to top-probability decoding, which is the default confidence-based decoding strategy considered in this work. To examine whether the method also applies to a different confidence definition, we additionally evaluate it with top-margin decoding. Top-margin decoding defines position confidence as the gap between the two largest predicted token probabilities, rather than the maximum probability itself. Table~\ref{tab:confidence_strategy_results} reports results with both top-probability and top-margin decoding on four text-only reasoning benchmarks. Across both LLaDA and LLaDA-1.5, suffix anchoring improves the corresponding top-margin baseline, and adding confidence modulation provides further gains. The full method improves the average accuracy from 23.98 to 51.07 for LLaDA and from 34.60 to 51.18 for LLaDA-1.5. These results show that our method is not limited to top-probability decoding and remains effective when applied to top-margin decoding.

\subsection{Code-Generation Results}
\label{app:code_generation}

\begin{table}[h]
\centering
\setlength{\tabcolsep}{3.5pt}
\resizebox{\linewidth}{!}{%
\begin{tabular}{ll|cc|c}
\toprule
\multicolumn{1}{l}{Model}
& \multicolumn{1}{l|}{Decoding Method}
& \multicolumn{1}{c}{HumanEval}
& \multicolumn{1}{c|}{MBPP}
& \multicolumn{1}{c}{Average} \\
\midrule
\midrule

\multirow{5}{*}{LLaDA}
& Uniform
& 18.90 & 22.40 & 20.65 \\
& Top Probability
& 17.07 & 19.20 & 18.14 \\
& Top Margin
& 17.07 & 24.40 & 20.74 \\
& Ours w/o Confidence Modulation
& 28.66 & 25.80 & 27.23 \\
& Ours
& \textbf{32.93} & \textbf{28.40} & \textbf{30.67} \\

\midrule

\multirow{5}{*}{LLaDA-1.5}
& Uniform
& 20.12 & 22.80 & 21.46 \\
& Top Probability
& 20.12 & 25.80 & 22.96 \\
& Top Margin
& 22.56 & 28.20 & 25.38 \\
& Ours w/o Confidence Modulation
& 29.88 & 28.00 & 28.94 \\
& Ours
& \textbf{35.98} & \textbf{30.60} & \textbf{33.29} \\

\bottomrule
\end{tabular}%
}
\caption{\textbf{Results on code-generation benchmarks.}
Pass@1~(\%) is reported on HumanEval and MBPP using LLaDA and LLaDA-1.5. We compare random position selection~(Uniform), confidence-based decoding baselines, our method without confidence modulation, and the full method. \textbf{Bold} indicates the best result for each model.}
\label{tab:code_generation_results}
\end{table}

% \begin{table}[h]
% \centering
% \setlength{\tabcolsep}{4pt}
% \resizebox{\linewidth}{!}{%
% \begin{tabular}{ll|cc|c}
% \toprule
% \multicolumn{1}{l}{Model}
% & \multicolumn{1}{l|}{Decoding Method}
% & \multicolumn{1}{c}{HumanEval}
% & \multicolumn{1}{c|}{MBPP}
% & \multicolumn{1}{c}{Average} \\
% \midrule
% \midrule
% \multirow{7}{*}{LLaDA}
% & Random                         & 18.90 & 22.40 & 20.65 \\
% \cmidrule(lr){2-5}
% & Top Probability                & 17.07 & 19.20 & 18.14 \\
% & \quad +  Suffix Anchor  & 28.66 & 25.80 & 27.23 \\
% & \quad + Confidence Modulation   & \textbf{32.93} & \textbf{28.40} & \textbf{30.67} \\
% \cmidrule(lr){2-5}
% & Top Margin                     & 17.07 & 24.40 & 20.74 \\
% & \quad +  Suffix Anchor  & 28.66 & 28.80 & 28.73 \\
% & \quad + Confidence Modulation   & \textbf{31.71} & \textbf{31.80} & \textbf{31.76} \\
% \bottomrule
% \end{tabular}%
% }
% \caption{\textbf{Results on code-generation benchmarks.} Pass@1~(\%) is reported on HumanEval and MBPP using LLaDA 8B-Instruct. For each confidence-based decoding strategy, the unmodified baseline, suffix anchoring, and the full method with confidence modulation are compared. Random position selection is included as a non-confidence-based reference. \textbf{Bold} indicates the best result within each confidence-based decoding group.}
% \label{tab:code_generation_results}
% \end{table}

Table~\ref{tab:code_generation_results} reports code-generation results on HumanEval and MBPP using LLaDA 8B-Instruct and LLaDA-1.5. Our method achieves the best performance for both models, with average pass@1 scores of $30.67$ and $33.29$, respectively. Suffix anchoring alone already improves over the decoding baselines, while confidence modulation further increases the average score from $27.23$ to $30.67$ on LLaDA and from $28.94$ to $33.29$ on LLaDA-1.5. These results show that the method also generalizes to code generation, where confidence modulation helps prevent premature selection of anchor-adjacent code positions.

\subsection{Ablation Over Suffix Anchors}
\label{app:suffix_anchor_choice}
\begin{table*}[h]
\centering
\setlength{\tabcolsep}{9pt}
\resizebox{0.75\linewidth}{!}{%
\begin{tabular}{l|l|l|c}
\toprule
Suffix Anchor & Model & Decoding Method & Accuracy \\
\midrule
\midrule

\multirow{2}{*}{No Suffix Anchor}
& LLaDA     & Top Probability & 14.94 \\
\cmidrule(lr){2-4}
& LLaDA-1.5 & Top Probability & 23.35 \\

\midrule

\multirow{4}{*}{\textit{``The answer is''} (Default)}
& \multirow{2}{*}{LLaDA}
& Ours w/o Confidence Modulation & 49.89 \\
&
& Ours                         & \textbf{76.88} \\
\cmidrule(lr){2-4}
& \multirow{2}{*}{LLaDA-1.5}
& Ours w/o Confidence Modulation & 51.40 \\
&
& Ours                         & \textbf{75.59} \\

\midrule

\multirow{4}{*}{\textit{``Therefore, the answer is''}}
& \multirow{2}{*}{LLaDA}
& Ours w/o Confidence Modulation & 52.77 \\
&
& Ours                         & \textbf{74.45} \\
\cmidrule(lr){2-4}
& \multirow{2}{*}{LLaDA-1.5}
& Ours w/o Confidence Modulation & 50.64 \\
&
& Ours                         & \textbf{73.62} \\

\midrule

\multirow{4}{*}{\textit{``Answer:''}}
& \multirow{2}{*}{LLaDA}
& Ours w/o Confidence Modulation & 49.58 \\
&
& Ours                         & \textbf{73.16} \\
\cmidrule(lr){2-4}
& \multirow{2}{*}{LLaDA-1.5}
& Ours w/o Confidence Modulation & 49.51 \\
&
& Ours                         & \textbf{73.24} \\

\midrule

\multirow{4}{*}{\textit{``,''}}
& \multirow{2}{*}{LLaDA}
& Ours w/o Confidence Modulation & 55.12 \\
&
& Ours                         & \textbf{73.77} \\
\cmidrule(lr){2-4}
& \multirow{2}{*}{LLaDA-1.5}
& Ours w/o Confidence Modulation & 51.86 \\
&
& Ours                         & \textbf{72.18} \\

\midrule

\multirow{4}{*}{\textit{``.''}}
& \multirow{2}{*}{LLaDA}
& Ours w/o Confidence Modulation & 54.13 \\
&
& Ours                         & \textbf{74.68} \\
\cmidrule(lr){2-4}
& \multirow{2}{*}{LLaDA-1.5}
& Ours w/o Confidence Modulation & 53.53 \\
&
& Ours                         & \textbf{75.36} \\

\midrule

\multirow{4}{*}{\textit{``is''}}
& \multirow{2}{*}{LLaDA}
& Ours w/o Confidence Modulation & 55.12 \\
&
& Ours                         & \textbf{73.62} \\
\cmidrule(lr){2-4}
& \multirow{2}{*}{LLaDA-1.5}
& Ours w/o Confidence Modulation & 55.65 \\
&
& Ours                         & \textbf{73.77} \\

\bottomrule
\end{tabular}%
}
\caption{\textbf{Ablation over suffix anchors.}
Accuracy~(\%) is reported on GSM8K using LLaDA 8B-Instruct and LLaDA-1.5 with top-probability decoding. Each suffix anchor is inserted at the same anchor position before decoding begins. Ours w/o Confidence Modulation denotes suffix anchoring without confidence modulation. \textbf{Bold} indicates the better result within each model and suffix-anchor setting.}
\label{tab:suffix_anchor_ablation}
\end{table*}

% \begin{table*}[h]
% \centering
% \setlength{\tabcolsep}{9pt}
% \resizebox{0.65\linewidth}{!}{%
% \begin{tabular}{l|l|c}
% \toprule
% Suffix Anchor & Decoding Method & Accuracy \\
% \midrule
% \midrule
% No Suffix Anchor & Unmodified Top Probability & 14.94 \\
% \midrule
% \multirow{2}{*}{\textit{``The answer is''} (Default)} & Suffix Anchoring & 49.89 \\ & + Confidence Modulation & \textbf{76.88} \\ \midrule \multirow{2}{*}{\textit{``Therefore, the answer is''}}
% & Suffix Anchoring & 52.77 \\
% & + Confidence Modulation & \textbf{74.45} \\
% \midrule
% \multirow{2}{*}{\textit{``Answer:''}} & Suffix Anchoring & 49.58 \\ & + Confidence Modulation & \textbf{73.16} \\ 
% \midrule
% \multirow{2}{*}{\textit{``is''}} & Suffix Anchoring & 55.12 \\ & + Confidence Modulation & \textbf{73.62} \\ \midrule \multirow{2}{*}{\textit{``,''}}
% & Suffix Anchoring & 55.12 \\
% & + Confidence Modulation & \textbf{73.77} \\
% \midrule
% \multirow{2}{*}{\textit{``.''}}
% & Suffix Anchoring & 54.13 \\
% & + Confidence Modulation & \textbf{74.68} \\
% \bottomrule
% \end{tabular}%
% }
% \caption{\textbf{Ablation over suffix anchors.} Accuracy~(\%) is reported on GSM8K using LLaDA 8B-Instruct with top-probability decoding. Each suffix anchor is inserted at the same anchor position before decoding begins. \textbf{Bold} indicates the best result within each suffix-anchor group.}
% \label{tab:suffix_anchor_ablation}
% \end{table*}
We ablate the suffix-anchor choice on GSM8K using LLaDA 8B-Instruct and LLaDA-1.5 with top-probability decoding. While the main experiments use \textit{``The answer is''} for reasoning benchmarks, Table~\ref{tab:suffix_anchor_ablation} evaluates several alternatives. Across both models, all tested anchors substantially improve over decoding without a suffix anchor, and adding confidence modulation consistently provides further gains. Notably, even the anchor \textit{``.''}, which provides minimal response structure, achieves $74.68$ on LLaDA and $75.36$ on LLaDA-1.5 with the full method, closely matching the default anchor. These results indicate that our suffix anchoring approach does not require a detailed response template; even a lightweight continuation cue can discourage premature EOT prediction. This distinguishes our approach from prior in-place prompting strategies that prescribe more explicit response structures or output constraints~\citep{xiong2025unveiling, jin2025thinking, lee2025unlocking}.

\subsection{Ablation Over Anchor Positions}
\label{app:anchor_position}
\begin{table*}[h]
\centering
\setlength{\tabcolsep}{7pt}
\begin{subtable}[t]{0.49\linewidth}
\centering
\resizebox{\linewidth}{!}{%
\begin{tabular}{l|l|cc}
\toprule
Anchor Position & Decoding Method & Acc. & EOT Ratio \\
\midrule
\midrule
No Anchor & Top Probability & 14.94 & 0.66 \\
\midrule
\multirow{2}{*}{$-20$ (Default)}
& Ours w/o CM & 49.89 & 0.05 \\
& Ours & 76.88 & 0.05 \\
\midrule
\multirow{2}{*}{$-30$}
& Ours w/o CM & 52.69 & 0.09 \\
& Ours & 76.88 & 0.09 \\
\midrule
\multirow{2}{*}{$-40$}
& Ours w/o CM & 55.04 & 0.13 \\
& Ours & 76.65 & 0.13 \\
\midrule
\multirow{2}{*}{$-50$}
& Ours w/o CM & 56.33 & 0.17 \\
& Ours & 77.33 & 0.17 \\
\midrule
\multirow{2}{*}{$-60$}
& Ours w/o CM & 56.48 & 0.21 \\
& Ours & 74.68 & 0.21 \\
\bottomrule
\end{tabular}
}
\caption{Suffix Anchor: \textit{''The answer is''} (Default)} \label{tab:anchor_position_answer_is} \end{subtable} \hfill \begin{subtable}[t]{0.49\linewidth} \centering \resizebox{\linewidth}{!}{ \begin{tabular}{l|l|cc} \toprule Anchor Position & Decoding Method & Acc. & EOT Ratio \\ \midrule \midrule No Anchor & Top Probability & 14.94 & 0.66 \\ \midrule \multirow{2}{*}{$-20$ (Default)} & Ours w/o CM & 54.13 & 0.07 \\ & Ours & 74.68 & 0.08 \\ \midrule \multirow{2}{*}{$-30$} & Ours w/o CM & 56.18 & 0.11 \\ & Ours & 75.06 & 0.12 \\ \midrule \multirow{2}{*}{$-40$} & Ours w/o CM & 56.48 & 0.15 \\ & Ours & 74.37 & 0.15 \\ \midrule \multirow{2}{*}{$-50$} & Ours w/o CM & 55.72 & 0.19 \\ & Ours & 74.91 & 0.19 \\ \midrule \multirow{2}{*}{$-60$} & Ours w/o CM & 56.94 & 0.23 \\ & Ours & 73.31 & 0.23 \\ \bottomrule \end{tabular} } \caption{Suffix Anchor: \textit{``.''}}
\label{tab:anchor_position_period}
\end{subtable}
\caption{\textbf{Ablation over anchor positions.} Accuracy~(\%) and EOT ratio are reported on GSM8K using LLaDA 8B-Instruct with top-probability decoding and generation length $L=256$. Anchor position $-k$ denotes inserting the suffix anchor $k$ positions before the end of the response region. Ours w/o CM refers to our method without confidence modulation.}
\label{tab:anchor_position_ablation}
\end{table*}
We ablate anchor positions on GSM8K using LLaDA 8B-Instruct with top-probability decoding and response length $L=256$. Table~\ref{tab:anchor_position_ablation} reports results for two suffix anchors, \textit{``The answer is''} and the semantically minimal anchor \textit{``.''}, while varying their insertion positions within the later response region. The default position is $-20$, meaning 20 positions before the end of the response region, and the ablation shifts the anchor earlier relative to this default position. For both suffix anchors, the full method maintains robust performance under these positional changes. At the same time, moving the anchor earlier generally increases the EOT ratio in generated outputs. This position-dependent change in EOT ratio supports our interpretation that the suffix anchor serves as a lightweight cue that conditions the model to generate non-EOT content up to positions near the anchor, rather than imposing a fixed response template.

\section{Qualitative Analysis}
\label{app:qualitative_analysis}

\subsection{Qualitative Comparison of Decoding Variants}
\label{app:qualitative_decoding_variants}
\begin{figure*}[t]
\begin{center}
\includegraphics[width=1.0\linewidth]{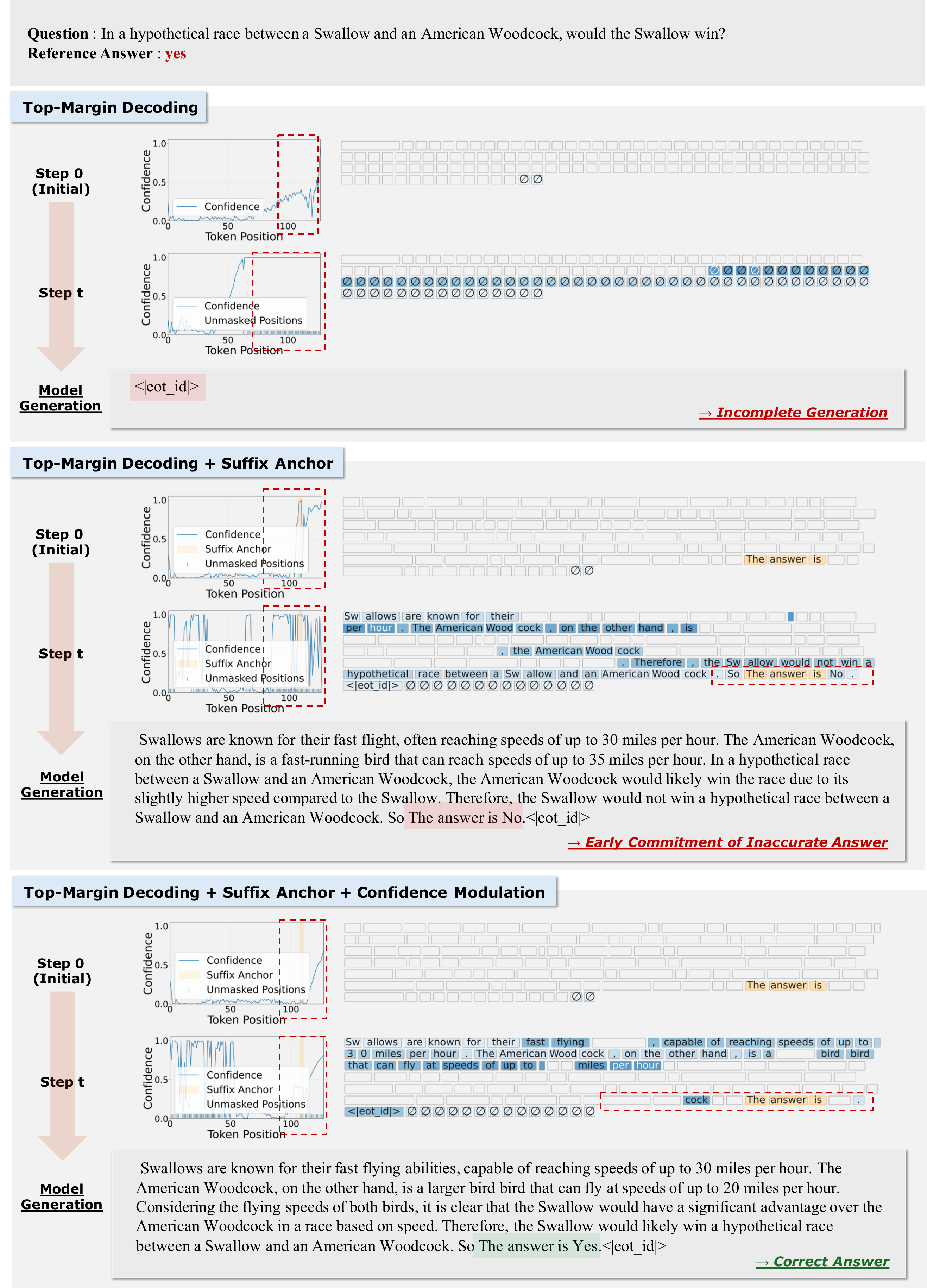}   
    \vspace{-6mm}
\end{center}
\caption{
\textbf{Qualitative example on StrategyQA under top-margin decoding.} The unmodified baseline, suffix anchoring, and the full method with confidence modulation are compared using LLaDA. It shows confidence over token positions and unmasked tokens at the initial step and an intermediate decoding step, along with the final output. Darker blue token boxes indicate positions decoded at later steps. $\varnothing$ denotes the \texttt{<|endoftext|>} token.
}
\vspace{-4mm}
\label{fig:StrategyQA_top-margin}
\end{figure*}

Figures~\ref{fig:GSM8K_top-prob}--\ref{fig:StrategyQA_top-margin} provide qualitative comparisons among the unmodified confidence-based baseline, suffix anchoring, and the full method with confidence modulation. For each figure, the subfigures corresponding to these three decoding variants visualize confidence over token positions~(left) and unmasked tokens~(right) at the initial step and an intermediate decoding step, together with the final output~(bottom). These examples illustrate how suffix anchoring mitigates incomplete generation, while confidence modulation reduces premature inaccurate decoding near the suffix anchor.

\subsection{Decoding Progress and Confidence Dynamics}
\label{app:decoding_progress}
\begin{figure*}[t]
\begin{center}
\includegraphics[width=1.0\linewidth]{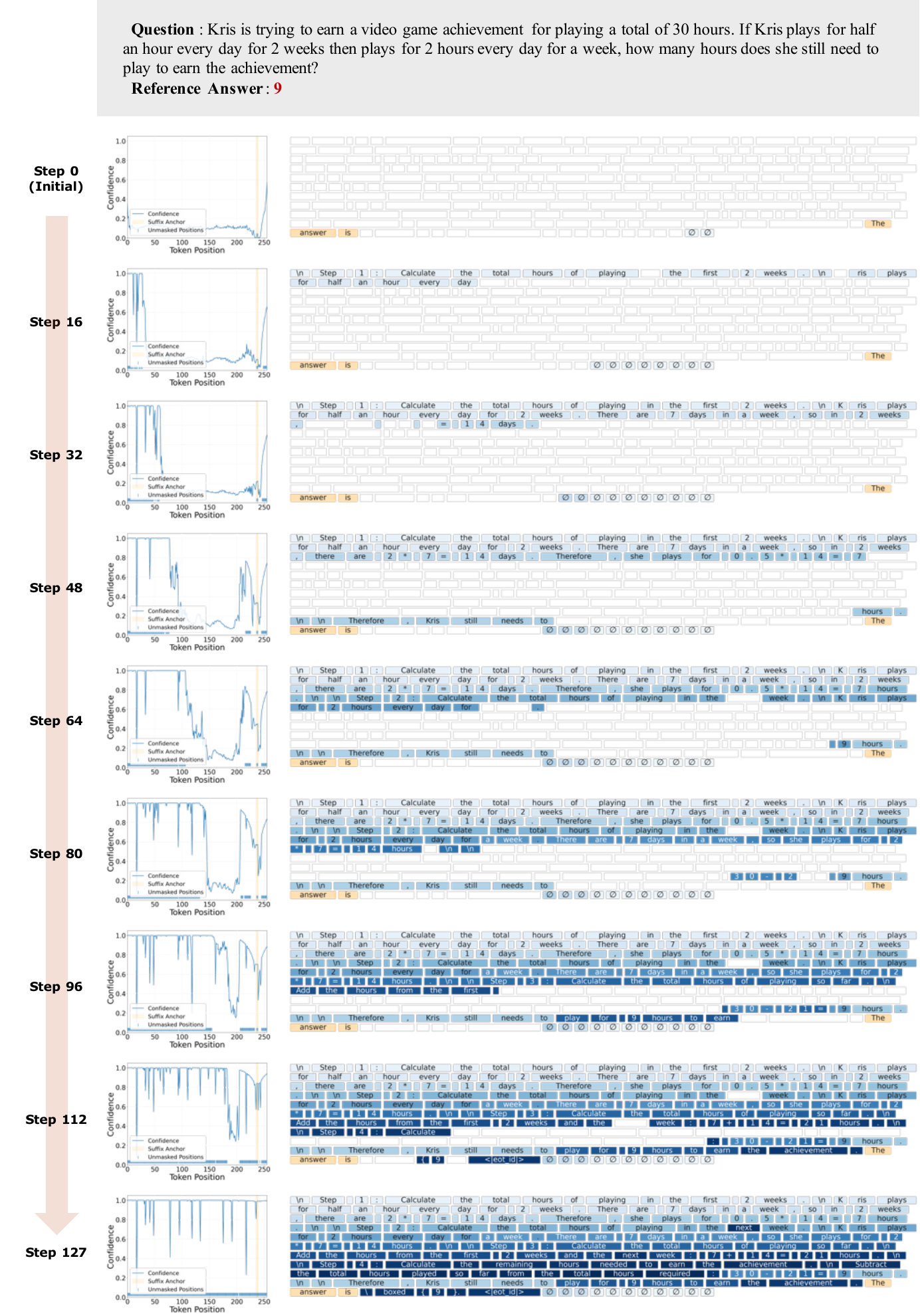}   
    \vspace{-6mm}
\end{center}
\caption{\textbf{Decoding progress of Suffix-Anchored Confidence Modulation.} Confidence over token positions~(left) and unmasked tokens~(right) are visualized from the initial step to the final decoding step for a GSM8K example using LLaDA under top-probability decoding. Darker blue token boxes indicate positions decoded at later steps. $\varnothing$ denotes the \texttt{<|endoftext|>} token.}
\vspace{-4mm}
\label{fig:Decoding progress}
\end{figure*}

Figure~\ref{fig:Decoding progress} visualizes the decoding process of Suffix-Anchored Confidence Modulation on a GSM8K example using LLaDA 8B-Instruct with top-probability decoding. The figure shows confidence over token positions and the corresponding unmasked tokens from the initial step to the final decoding step. This illustrates how the method gradually resolves the response while progressively relaxing the confidence modulation near the suffix anchor, allowing anchor-adjacent positions to be decoded after more surrounding context has been generated.

\section{Use of LLMs in This Work} 
LLM-based assistance was used during the preparation of this work. Specifically, LLMs were used to support code implementation and debugging, and to improve the clarity, grammar, and readability of the manuscript. All scientific ideas, methodological decisions, experimental analyses, and interpretations originated from the authors. Any code or text produced with LLM assistance was carefully reviewed, verified, and edited by the authors before being included in this work. The authors take full responsibility for the content of the manuscript.

\begin{figure*}[t]
\begin{center}
\includegraphics[width=0.9\linewidth]{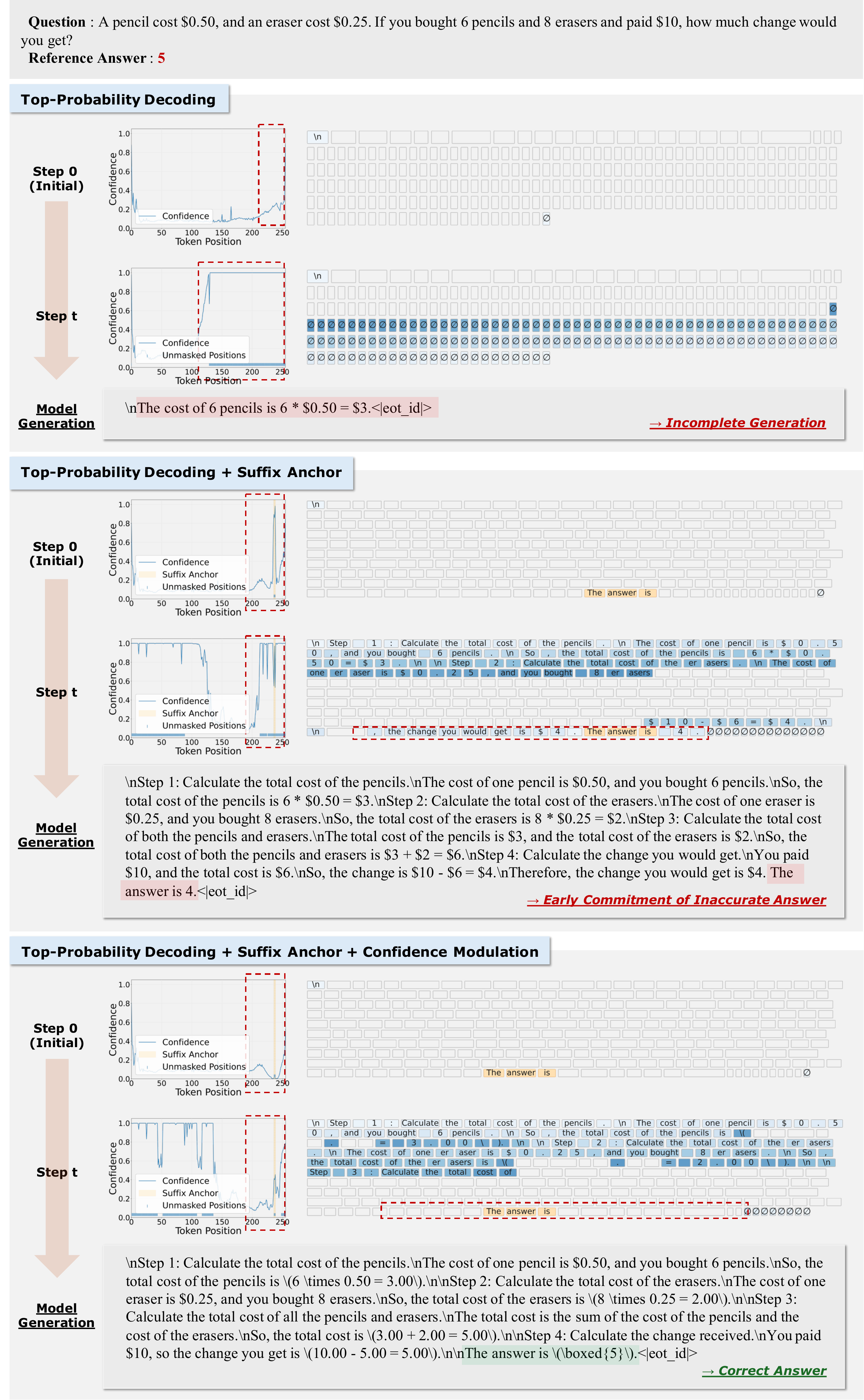}   
    \vspace{-3mm}
\end{center}
\caption{\textbf{Qualitative example on GSM8K under top-probability decoding.} The unmodified baseline, suffix anchoring, and the full method with confidence modulation are compared using LLaDA. It shows confidence over token positions and unmasked tokens at the initial step and an intermediate decoding step, along with the final output. Darker blue token boxes indicate positions decoded at later steps. $\varnothing$ denotes the \texttt{<|endoftext|>} token.}
\vspace{-4mm}
\label{fig:GSM8K_top-prob}
\end{figure*}

\begin{figure*}[t]
\begin{center}
\includegraphics[width=0.9\linewidth]{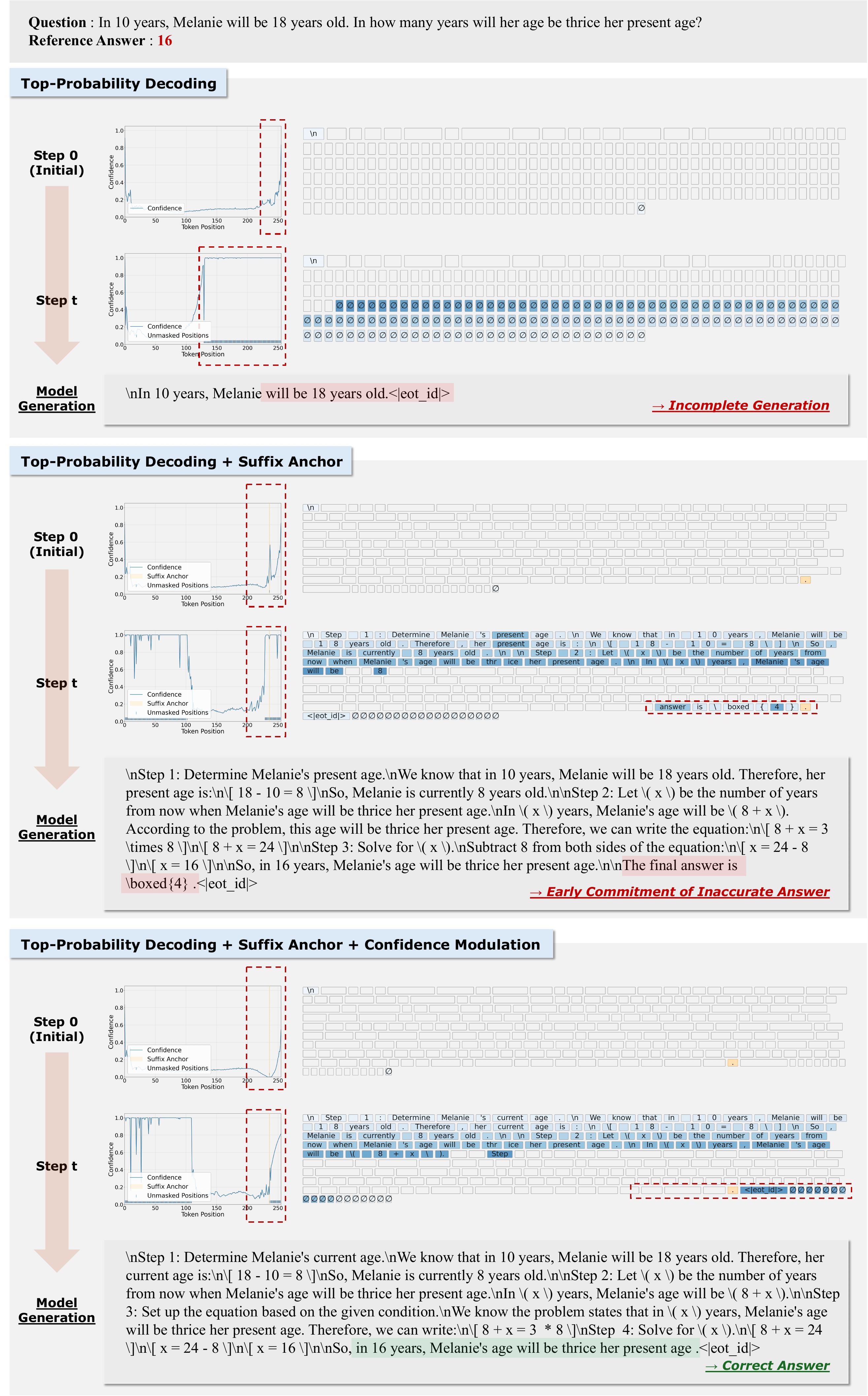}   
    \vspace{-3mm}
\end{center}
\caption{\textbf{Qualitative example on GSM8K under top-probability decoding.} The unmodified baseline, suffix anchoring, and the full method with confidence modulation are compared using LLaDA. We use ``.'' as the suffix anchor. It shows confidence over token positions and unmasked tokens at the initial step and an intermediate decoding step, along with the final output. Darker blue token boxes indicate positions decoded at later steps. $\varnothing$ denotes the \texttt{<|endoftext|>} token.}
\vspace{-4mm}
\label{fig:GSM8K_top-prob}
\end{figure*}

\begin{figure*}[t]
\begin{center}
\includegraphics[width=0.89\linewidth]{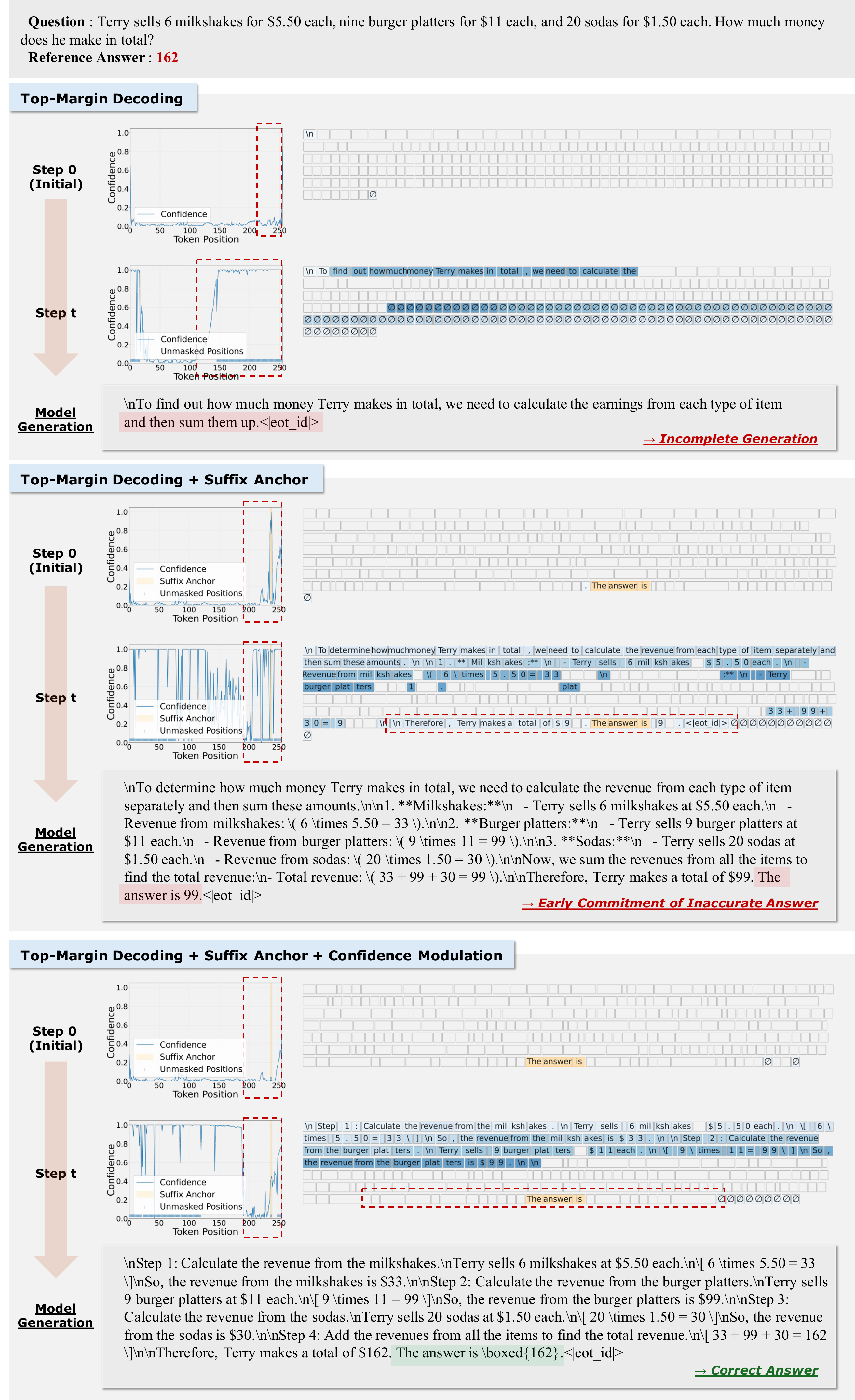}   
    \vspace{-3.5mm}
\end{center}
\caption{
\textbf{Qualitative example on GSM8K under top-margin decoding.} The unmodified baseline, suffix anchoring, and the full method with confidence modulation are compared using LLaDA. It shows confidence over token positions and unmasked tokens at the initial step and an intermediate decoding step, along with the final output. Darker blue token boxes indicate positions decoded at later steps. $\varnothing$ denotes the \texttt{<|endoftext|>} token.
}
\vspace{-4mm}
\label{fig:GSM8K_top-margin}
\end{figure*}

\begin{figure*}[t]
\begin{center}
\includegraphics[width=0.9\linewidth]{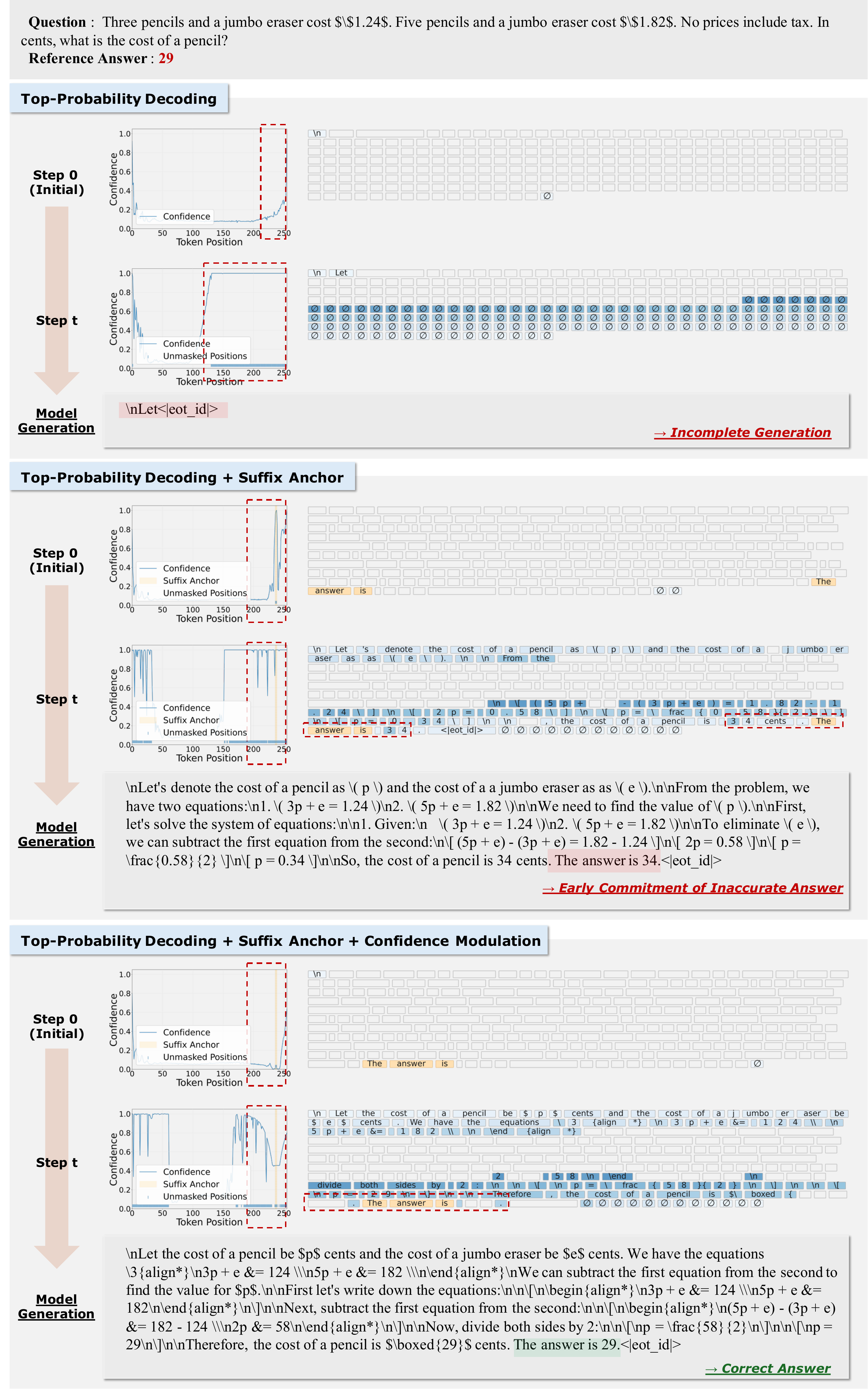}   
    \vspace{-3.5mm}
\end{center}
\caption{
\textbf{Qualitative example on MATH-500 under top-probability decoding.} The unmodified baseline, suffix anchoring, and the full method with confidence modulation are compared using LLaDA. It shows confidence over token positions and unmasked tokens at the initial step and an intermediate decoding step, along with the final output. Darker blue token boxes indicate positions decoded at later steps. $\varnothing$ denotes the \texttt{<|endoftext|>} token.
}
\vspace{-4mm}
\label{fig:MATH-500_top-prob}
\end{figure*}

\begin{figure*}[t]
\begin{center}
\includegraphics[width=0.9\linewidth]{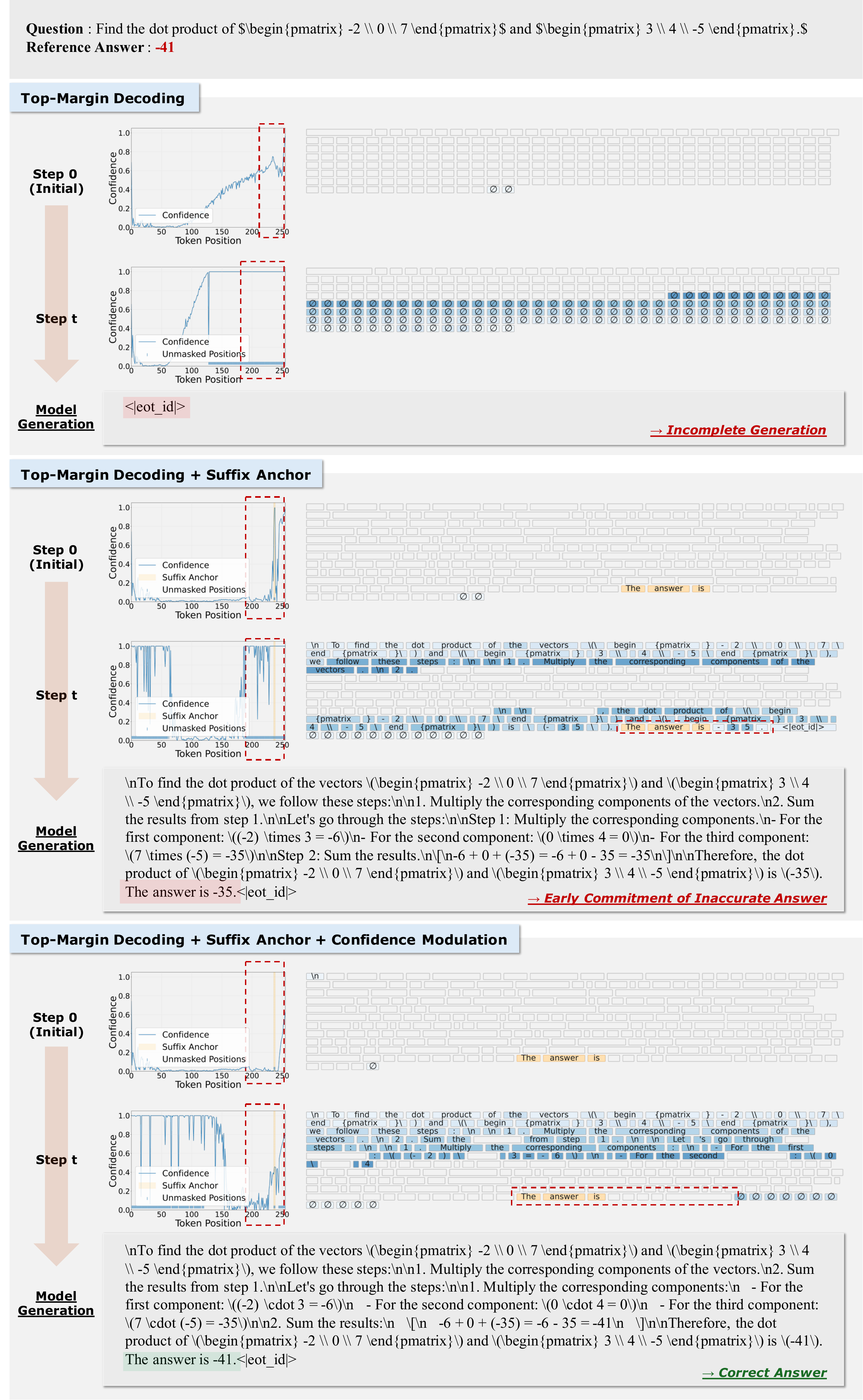}   
    \vspace{-3.5mm}
\end{center}
\caption{
\textbf{Qualitative example on MATH-500 under top-margin decoding.} The unmodified baseline, suffix anchoring, and the full method with confidence modulation are compared using LLaDA. It shows confidence over token positions and unmasked tokens at the initial step and an intermediate decoding step, along with the final output. Darker blue token boxes indicate positions decoded at later steps. $\varnothing$ denotes the \texttt{<|endoftext|>} token.
}
\vspace{-4mm}
\label{fig:MATH-500_top-margin}
\end{figure*}

\begin{figure*}[t]
\begin{center}
\includegraphics[width=1.0\linewidth]{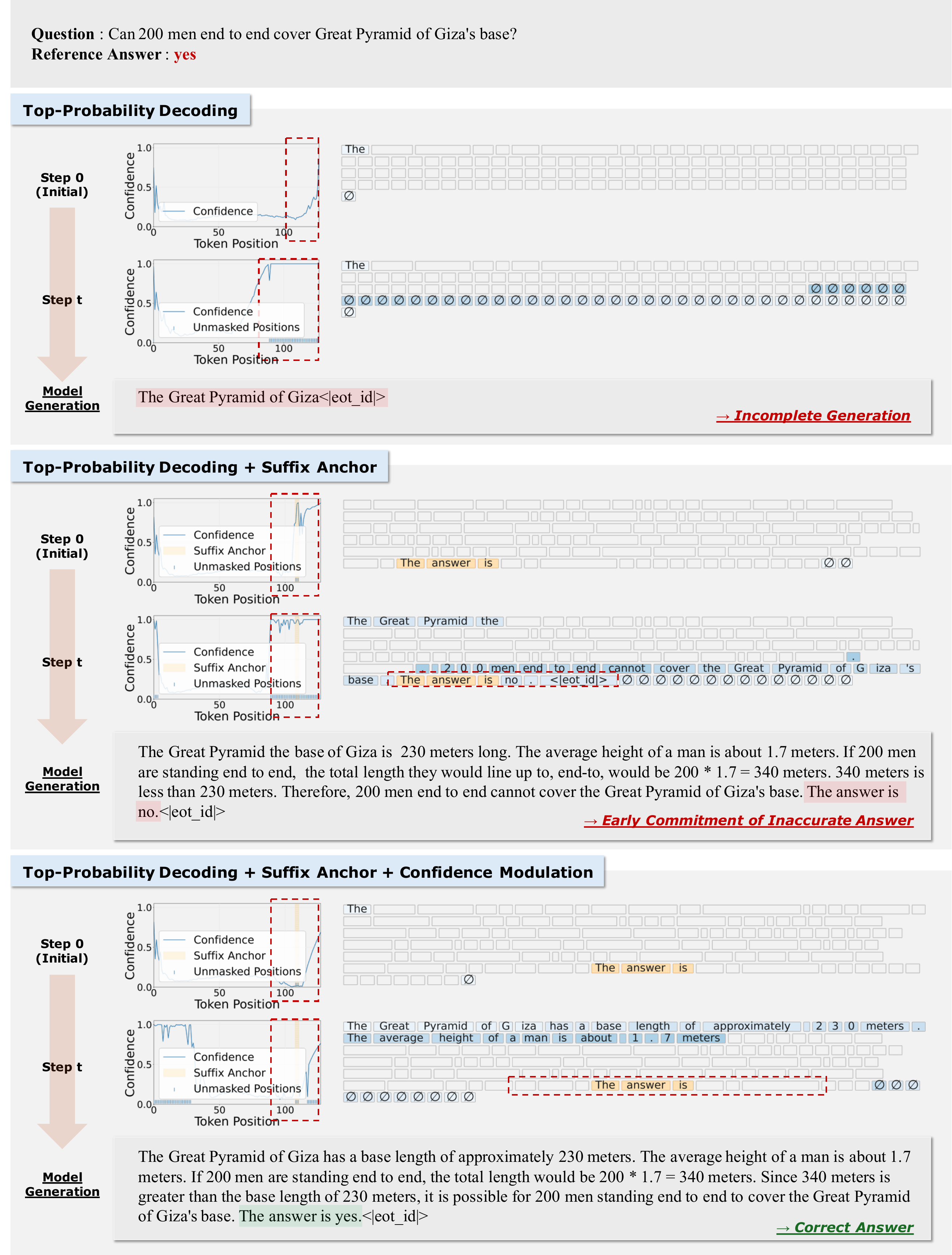}   
    \vspace{-6mm}
\end{center}
\caption{
\textbf{Qualitative example on StrategyQA under top-probability decoding.} The unmodified baseline, suffix anchoring, and the full method with confidence modulation are compared using LLaDA. It shows confidence over token positions and unmasked tokens at the initial step and an intermediate decoding step, along with the final output. Darker blue token boxes indicate positions decoded at later steps. $\varnothing$ denotes the \texttt{<|endoftext|>} token.
}
\vspace{-4mm}
\label{fig:StrategyQA_top-prob}
\end{figure*}

% \begin{figure*}[t]
% \begin{center}
% \includegraphics[width=0.98\linewidth]{Figures/appendix_mathvista_example.pdf}   
%     \vspace{-2mm}
% \end{center}
% \caption{
% \textbf{Qualitative example on MathVista under top-probability decoding.} The unmodified baseline, suffix anchoring, and the full method with confidence modulation are compared using LaViDa. It shows confidence over token positions and unmasked tokens at the initial step and an intermediate decoding step, along with the final output. Darker blue token boxes indicate positions decoded at later steps. $\varnothing$ denotes the \texttt{<|endoftext|>} token.
% }
% \vspace{-4mm}
% \label{fig:MathVista_top-prob}
% \end{figure*}

% \begin{figure*}[t]
% \begin{center}
% \includegraphics[width=0.98\linewidth]{Figures/appendix_mathvista_topmargin_example.pdf}   
%     \vspace{-2mm}
% \end{center}
% \caption{
% \textbf{Qualitative example on MathVista under top-margin decoding.} The unmodified baseline, suffix anchoring, and the full method with confidence modulation are compared using LaViDa. It shows confidence over token positions and unmasked tokens at the initial step and an intermediate decoding step, along with the final output. Darker blue token boxes indicate positions decoded at later steps. $\varnothing$ denotes the \texttt{<|endoftext|>} token.
% }
% \vspace{-4mm}
% \label{fig:MathVista_top-margin}
% \end{figure*}

\clearpage

\begin{table*}[h]
    \centering
    \small

    \begin{tabular}{
        @{}
        >{\raggedright\arraybackslash\bfseries}p{3cm}
        >{\raggedright\arraybackslash}p{0.76\textwidth}
        @{}
    }
    \toprule
    \textbf{Benchmark} & \textbf{Prompt Template} \\
    \midrule

    GSM8K (0-shot)
    &
    Q: \{question\}\newline
    A: Let's think step by step.
    \\
    \noalign{\vskip 0.5em}
    \midrule

    MATH500 (0-shot)
    &
    Q: \{question\}\newline
    A: Let's think step by step. End your solution with a final line in
    the exact format ``The answer is <final answer>'' and do not write
    anything after that final answer line.
    \\
    \noalign{\vskip 0.5em}
    \midrule

    StrategyQA (0-shot)
    &
    Q: \{question\}\newline
    A: Let's think step by step.
    \\
    \noalign{\vskip 0.5em}
    \midrule

    MMLU-Pro (5-shot)
    &
    The following are multiple choice questions (with answers) about
    \{subject\}.\newline
    Think step by step and then finish your answer with
    ``the answer is (X),'' where X is the correct letter choice.\newline
    \{fewshot\_examples\}\newline
    Question:\newline
    \{question\}\newline
    Options:\newline
    \{choices\}\newline
    Answer: Let's think step by step.
    \\
    \noalign{\vskip 0.5em}
    \midrule

    MathVista (0-shot)\newline
    (multiple-choice)
    &
    <image>\newline
    \{question\}\newline\newline
    \{choices\}\newline\newline
    Hint: First perform reasoning, then provide the correct option letter,
    e.g., A, B, C, or D, at the end in the format:
    The answer is <final answer>.
    \\
    \noalign{\vskip 0.5em}
    \midrule

    MathVista (0-shot)\newline
    (free-form)
    &
    <image>\newline
    \{question\}\newline\newline
    Hint: First perform reasoning, then provide the final
    \{answer\_type\_hint\} at the end in the format:
    The answer is <final answer>.
    \\
    \noalign{\vskip 0.5em}
    \midrule

    ChartQA (0-shot)
    &
    <image>\newline
    \{question\}\newline
    Let's think step by step. Then conclude on the last line with:
    The answer is <final answer>.
    \\
    \noalign{\vskip 0.5em}
    \midrule

    HumanEval (0-shot)
    &
    \{question\}
    \\
    \noalign{\vskip 0.5em}
    \midrule

    MBPP (3-shot)
    &
    \{fewshot\_examples\}\newline
    You are an expert Python programmer, and here is your task:
    \{question\} Your code should pass these tests:\newline\newline
    \{test\_1\}\newline
    \{test\_2\}\newline
    \{test\_3\}\newline
    [BEGIN]
    \\

    \bottomrule
    \end{tabular}

    \caption{\textbf{Prompt templates used for evaluation.} The table reports the benchmark-specific prompts and shot settings used in our experiments. Text enclosed in braces denotes a placeholder replaced with the corresponding benchmark instance, choices, few-shot examples, or test cases.}
    \label{tab:prompt_template}
\end{table*}

\end{document}